\PassOptionsToPackage{unicode}{hyperref}
\PassOptionsToPackage{hyphens}{url}
\PassOptionsToPackage{dvipsnames,svgnames,x11names}{xcolor}
\documentclass[
]{article}
\usepackage{xcolor}
\usepackage{amsmath,amssymb}
\usepackage{iftex}
\ifPDFTeX
  \usepackage[T1]{fontenc}
  \usepackage[utf8]{inputenc}
  \usepackage{textcomp} 
\else 
  \usepackage{unicode-math} 
  \defaultfontfeatures{Scale=MatchLowercase}
  \defaultfontfeatures[\rmfamily]{Ligatures=TeX,Scale=1}
\fi
\usepackage{lmodern}
\ifPDFTeX\else
\fi
\IfFileExists{upquote.sty}{\usepackage{upquote}}{}
\IfFileExists{microtype.sty}{
  \usepackage[]{microtype}
  \UseMicrotypeSet[protrusion]{basicmath} 
}{}
\makeatletter
\@ifundefined{KOMAClassName}{
  \IfFileExists{parskip.sty}{%
    \usepackage{parskip}
  }{
    \setlength{\parindent}{0pt}
    \setlength{\parskip}{6pt plus 2pt minus 1pt}}
}{
  \KOMAoptions{parskip=half}}
\makeatother
\makeatletter
\ifx\paragraph\undefined\else
  \let\oldparagraph\paragraph
  \renewcommand{\paragraph}{
    \@ifstar
      \xxxParagraphStar
      \xxxParagraphNoStar
  }
  \newcommand{\xxxParagraphStar}[1]{\oldparagraph*{#1}\mbox{}}
  \newcommand{\xxxParagraphNoStar}[1]{\oldparagraph{#1}\mbox{}}
\fi
\ifx\subparagraph\undefined\else
  \let\oldsubparagraph\subparagraph
  \renewcommand{\subparagraph}{
    \@ifstar
      \xxxSubParagraphStar
      \xxxSubParagraphNoStar
  }
  \newcommand{\xxxSubParagraphStar}[1]{\oldsubparagraph*{#1}\mbox{}}
  \newcommand{\xxxSubParagraphNoStar}[1]{\oldsubparagraph{#1}\mbox{}}
\fi
\makeatother

\usepackage{color}
\usepackage{fancyvrb}

\DefineVerbatimEnvironment{Highlighting}{Verbatim}{commandchars=\\\{\}}
\usepackage{framed}
\definecolor{shadecolor}{RGB}{241,243,245}
\newenvironment{Shaded}{\begin{snugshade}}{\end{snugshade}}

\newcommand{\AttributeTok}[1]{\textcolor[rgb]{0.40,0.45,0.13}{#1}}

\newcommand{\CommentTok}[1]{\textcolor[rgb]{0.37,0.37,0.37}{#1}}

\newcommand{\ConstantTok}[1]{\textcolor[rgb]{0.56,0.35,0.01}{#1}}
\newcommand{\ControlFlowTok}[1]{\textcolor[rgb]{0.00,0.23,0.31}{\textbf{#1}}}

\newcommand{\DecValTok}[1]{\textcolor[rgb]{0.68,0.00,0.00}{#1}}

\newcommand{\FloatTok}[1]{\textcolor[rgb]{0.68,0.00,0.00}{#1}}
\newcommand{\FunctionTok}[1]{\textcolor[rgb]{0.28,0.35,0.67}{#1}}

\newcommand{\NormalTok}[1]{\textcolor[rgb]{0.00,0.23,0.31}{#1}}

\newcommand{\OtherTok}[1]{\textcolor[rgb]{0.00,0.23,0.31}{#1}}

\newcommand{\SpecialCharTok}[1]{\textcolor[rgb]{0.37,0.37,0.37}{#1}}

\newcommand{\StringTok}[1]{\textcolor[rgb]{0.13,0.47,0.30}{#1}}

\usepackage{longtable,booktabs,array}
\usepackage{calc} 
\usepackage{etoolbox}
\makeatletter
\patchcmd\longtable{\par}{\if@noskipsec\mbox{}\fi\par}{}{}
\makeatother
\IfFileExists{footnotehyper.sty}{\usepackage{footnotehyper}}{\usepackage{footnote}}
\makesavenoteenv{longtable}
\usepackage{graphicx}
\makeatletter
\newsavebox\pandoc@box
\newcommand*\pandocbounded[1]{
  \sbox\pandoc@box{#1}%
  \Gscale@div\@tempa{\textheight}{\dimexpr\ht\pandoc@box+\dp\pandoc@box\relax}%
  \Gscale@div\@tempb{\linewidth}{\wd\pandoc@box}%
  \ifdim\@tempb\p@<\@tempa\p@\let\@tempa\@tempb\fi
  \ifdim\@tempa\p@<\p@\scalebox{\@tempa}{\usebox\pandoc@box}%
  \else\usebox{\pandoc@box}%
  \fi%
}
\def\fps@figure{htbp}
\makeatother

\NewDocumentCommand\citeproctext{}{}
\NewDocumentCommand\citeproc{mm}{%
  \begingroup\def\citeproctext{#2}\cite{#1}\endgroup}
\makeatletter
 \let\@cite@ofmt\@firstofone
 \def\@biblabel#1{}
 \def\@cite#1#2{{#1\if@tempswa , #2\fi}}
\makeatother
\newlength{\cslhangindent}
\newlength{\csllabelwidth}
\newenvironment{CSLReferences}[2] 
 {\begin{list}{}{%
  \setlength{\itemindent}{0pt}
  \setlength{\leftmargin}{0pt}
  \setlength{\parsep}{0pt}
  \ifodd #1
   \setlength{\leftmargin}{\cslhangindent}
   \setlength{\itemindent}{-1\cslhangindent}
  \fi
  \setlength{\itemsep}{#2\baselineskip}}}
 {\end{list}}
\usepackage{calc}

\providecommand{\tightlist}{%
  \setlength{\itemsep}{0pt}\setlength{\parskip}{0pt}}

\usepackage{booktabs}
\usepackage{longtable}
\usepackage{array}
\usepackage{multirow}
\usepackage{wrapfig}
\usepackage{float}
\usepackage{colortbl}
\usepackage{pdflscape}
\usepackage{tabularx}
\usepackage{xltabular}
\usepackage{threeparttable}
\usepackage{threeparttablex}
\usepackage[normalem]{ulem}
\usepackage{makecell}
\usepackage{xcolor}

\newcommand{\pobs}{p_{\text{obs}}}
\newcommand{\psyn}{p_{\text{syn}}}
\newcommand{\nobs}{n_{\text{obs}}}
\newcommand{\nsyn}{n_{\text{syn}}}
\newcommand{\bx}{\mathbf{x}}
\usepackage{arxiv}
\usepackage{orcidlink}
\usepackage{amsmath}
\usepackage[T1]{fontenc}
\usepackage{float}
\makeatletter
\@ifpackageloaded{caption}{}{\usepackage{caption}}
\AtBeginDocument{%
\ifdefined\contentsname
  \renewcommand*\contentsname{Table of contents}
\else
  \newcommand\contentsname{Table of contents}
\fi
\ifdefined\listfigurename
  \renewcommand*\listfigurename{List of Figures}
\else
  \newcommand\listfigurename{List of Figures}
\fi
\ifdefined\listtablename
  \renewcommand*\listtablename{List of Tables}
\else
  \newcommand\listtablename{List of Tables}
\fi
\ifdefined\figurename
  \renewcommand*\figurename{Figure}
\else
  \newcommand\figurename{Figure}
\fi
\ifdefined\tablename
  \renewcommand*\tablename{Table}
\else
  \newcommand\tablename{Table}
\fi
}
\@ifpackageloaded{float}{}{\usepackage{float}}
\floatstyle{ruled}
\@ifundefined{c@chapter}{\newfloat{codelisting}{h}{lop}}{\newfloat{codelisting}{h}{lop}[chapter]}
\floatname{codelisting}{Listing}

\makeatother
\makeatletter
\@ifpackageloaded{caption}{}{\usepackage{caption}}
\@ifpackageloaded{subcaption}{}{\usepackage{subcaption}}
\makeatother
\usepackage{bookmark}
\IfFileExists{xurl.sty}{\usepackage{xurl}}{} 
\hypersetup{
  pdftitle={A density ratio framework for evaluating the utility of synthetic data},
  pdfauthor={Thom Benjamin Volker; Peter-Paul de Wolf; Erik-Jan van Kesteren},
  pdfkeywords={density ratio, disclosure limitation, privacy, synthetic
data, utility},
  colorlinks=true,
  linkcolor={blue},
  filecolor={Maroon},
  citecolor={Blue},
  urlcolor={Blue},
  pdfcreator={LaTeX via pandoc}}

\newcommand{\runninghead}{A Preprint }
\renewcommand{\runninghead}{Density ratios for utility }
\title{A density ratio framework for evaluating the utility of synthetic
data}
\def\asep{\\\\\\ } 
\def\asep{\And }
\author{\textbf{Thom Benjamin
Volker}~\orcidlink{0000-0002-2408-7820}\\Methodology and Statistics
\textbar{} Methodology\\Utrecht University \textbar{} Statistics
Netherlands\\Utrecht,\ 3584CH\\\href{mailto:t.b.volker@uu.nl}{t.b.volker@uu.nl}\asep\textbf{Peter-Paul
de Wolf}\\Methodology\\Statistics Netherlands\\The
Hague,\ 2490HA\\\href{mailto:pp.dewolf@cbs.nl}{pp.dewolf@cbs.nl}\asep\textbf{Erik-Jan
van Kesteren}\\Methodology and Statistics\\Utrecht
University\\Utrecht,\ 3584CH\\\href{mailto:e.vankesteren1@uu.nl}{e.vankesteren1@uu.nl}}
\date{}
\begin{document}
\maketitle
\begin{abstract}
Synthetic data generation is a promising technique to facilitate the use
of sensitive data while mitigating the risk of privacy breaches. For
synthetic data to be useful in downstream tasks, it needs to be of
sufficient quality. Various methods have been proposed to measure the
utility of synthetic data, but their results are often incomplete or
even misleading. We propose to use the density ratio estimation
framework to improve the evaluation, and thereby the quality, of
synthetic data. We show how this framework relates to and builds on
existing measures, yielding global and local utility measures that are
informative and easy to interpret. The methodology employs automatic,
non-parametric model selection for the density ratio, and thus requires
little to no manual model specification. In multivariate simulations,
density ratio-based methods compare favorably to classical utility
measures, such as the propensity mean squared error and the maximum mean
discrepancy. A real-world data application demonstrates how the density
ratio gives rise to local utility estimates that can guide refinements
of synthesis models. We conclude that density ratio estimation is a
valuable tool in synthetic data generation workflows and make the
methodology available through the open source R-package
\texttt{densityratio}.
\end{abstract}
{\bfseries \emph Keywords}
\def\sep{\textbullet\ }
density ratio \sep disclosure limitation \sep privacy \sep synthetic
data \sep 
utility

\section{Introduction}\label{introduction}

\linespread{1}

\linespread{2}

Openly accessible research data accelerates scientific progress
tremendously. Open data allows third-party researchers to answer
research questions with already collected data, freeing up resources
that would otherwise be devoted to data collection
(\citeproc{ref-ramachandran_open_2021}{Ramachandran, Bugbee, and Murphy
2021}). Sharing data with the accompanying analysis code allows others
to validate research findings and build upon the work
(\citeproc{ref-obels_analysis_2020}{Obels et al. 2020};
\citeproc{ref-crosas_automating_2015}{Crosas et al. 2015}). Students can
benefit from open data, as it fosters education with realistic data
(\citeproc{ref-atenas_open_2015}{Atenas, Havemann, and Priego 2015}),
and so can the general public, through stimulating citizen science
projects (\citeproc{ref-newman_future_2012}{Newman et al. 2012}).
However, making data openly available is (rightfully) hampered by
official legislation, like the General Data Protection Regulation (GDPR;
\citeproc{ref-gdpr}{European Parliament and Council of the European
Union 2016}), and general privacy concerns. In the worst case, sharing
data may cause harm to individuals or organizations, which may withhold
these entities from participating in future research. These privacy
constraints have been named among the biggest hurdles in the advancement
of computational social science (\citeproc{ref-lazer_css_2009}{Lazer et
al. 2009}), and among top reasons for companies to not share their data
with researchers (\citeproc{ref-fpf_2017}{Future of Privacy Forum
2017}).

Multiple approaches exist to balance the benefits of open data with
potential privacy risks. Traditionally, data providers employed a suite
of different disclosure limitation techniques before sharing the data,
such as top-coding, record-swapping or adding noise (e.g.,
\citeproc{ref-hundepool_disclosure_2012}{Hundepool et al. 2012};
\citeproc{ref-willenborg_elements_2001}{Willenborg and De Waal 2001}).
More recently, synthetic data has gained traction as a means to
disseminate private data (\citeproc{ref-SIPP_Beta_2006}{Abowd, Stinson,
and Benedetto 2006}; \citeproc{ref-hawala_synthetic_2008}{Hawala 2008};
\citeproc{ref-drechsler2012}{Drechsler 2012};
\citeproc{ref-vandewiel2023}{van de Wiel et al. 2023};
\citeproc{ref-obermeyer2019}{Obermeyer et al. 2019};
\citeproc{ref-zettler2021}{Zettler et al. 2021}), although the
conceptual framework traces back to the previous century
(\citeproc{ref-little_statistical_1993}{Little 1993};
\citeproc{ref-rubin_statistical_1993}{Rubin 1993}). Simply put, the idea
of synthetic data is to replace some, or all, of the observed values in
a data set by synthetic values that are generated from some model (e.g.,
\citeproc{ref-drechsler2011synthetic}{Drechsler 2011}). If only some
values are replaced, disclosure risks can be reduced because the
sensitive or identifying values do not correspond to their true values
anymore. If all values are replaced, there is also no one-to-one mapping
between the original and the synthetic data, further reducing the
disclosure risk. However, an increase in privacy typically comes at the
cost of a decrease in utility. As more of the data is altered, the
quality of the released data becomes more sensitive to the suitability
of the generative model. Regardless of the approach to disclosure
limitation, if the technique used to generate or alter the data does not
align with the intricacies of the problem at hand, the utility of the
released data will be further reduced than necessary.

Given that all disclosure limitation techniques reduce the utility of
the data, the challenge that arises is how to determine whether the
released data still has acceptable utility. Alternatively, one might
consider different disclosure limitation techniques that all satisfy the
defined privacy restrictions, and employ the one which yields data with
the highest utility. That is, given a set of methods that all meet the
privacy restrictions, one may aim to maximize the utility of the
released data. Both strategies require a reliable and encompassing
measure of data utility that allows to evaluate the quality of the
released data, and that allows to compare different disclosure
limitation techniques and/or synthesis models in terms of utility.
Moreover, adequate utility measures often guide the synthesis process,
by providing detailed feedback on important discrepancies between the
original and synthetic data. Lastly, good utility measures help the data
user in determining what the synthetic data can and cannot be used for.

In the synthetic data field, three classes of utility measures have been
distinguished (see \citeproc{ref-drechsler2023}{Drechsler and Haensch
2024} for a thorough review): fit-for-purpose measures,
analysis-specific utility measures and global utility measures.
Fit-for-purpose measures are often the first step in assessing the
quality of the synthetic data. These typically involve comparing the
univariate distributions of the observed and synthetic data (for example
using visualization techniques or goodness-of-fit measures). Although
these measures provide an initial impression of the quality of the
synthesis models used, this picture is by definition limited, because
only one or two variables are assessed at the same time. Hence, complex
relationships between variables will always be out of scope. Such
relationships may be captured by analysis-specific utility measures,
which quantify whether analyses on synthetic data provide results that
are comparable to results from the same analysis performed on the
observed data. These measures can, for example, evaluate how similar the
coefficients of a regression model are (e.g., using the confidence
interval overlap; \citeproc{ref-karr_utility_2006}{Karr et al. 2006}),
or whether prediction models trained on the synthetic and observed data
perform comparably in terms of evaluation metrics. However,
analysis-specific utility generally does not carry over: high specific
utility for one analysis does not at all imply high utility for another
analysis. Since data providers typically do not know which analyses will
be performed with the synthetic data, it is impossible to provide
analysis-specific utility measures for all potentially relevant analyses
(see also \citeproc{ref-drechsler_utility_2022}{Drechsler 2022}).

Global utility measures may overcome the shortcomings of the previous
approaches, as they evaluate the discrepancy between the entire
multivariate distribution of the observed and synthetic data. As such,
global utility measures are the most general, because if the observed
and synthetic data have similar (multivariate) distributions, all
potential analyses should yield similar results. Global utility can be
evaluated using some divergence measure, such as the Kullback-Leibler
divergence (\citeproc{ref-karr_utility_2006}{Karr et al. 2006}), or by
evaluating whether the observed and synthetic data are distinguishable
using a classification model (\citeproc{ref-Woo_global_2009}{Woo et al.
2009}; \citeproc{ref-snoke_utility_2018}{Snoke et al. 2018}). A commonly
used technique within the latter framework is the propensity Mean
Squared Error (\(pMSE\); \citeproc{ref-Woo_global_2009}{Woo et al.
2009}; \citeproc{ref-snoke_utility_2018}{Snoke et al. 2018}), which
quantifies how much the predicted propensity scores deviate from those
expected by chance. Alternative methods for quantifying global utility
are, for example, the discriminator loss in a generative adversarial
network (\citeproc{ref-goodfellow_gan_2014}{Goodfellow et al. 2014}) and
SPECKS (\citeproc{ref-Bowen_differentially_2021}{Bowen, Liu, and Su
2021}), which quantifies the distributional difference between two sets
of probabilities using the Kolmogorov-Smirnov distance. However, a
common critique of global utility measures is that they tend to be too
general (\citeproc{ref-drechsler_utility_2022}{Drechsler 2022}). That
is, analyses on a synthetic data set that is overall quite similar to
the observed data (i.e., has high global utility), may still yield
results that are far from the results obtained from the real data.
Alternatively, synthetic data with poor global utility might still
perform better for some tasks than alternative synthetic data sets with
higher global utility. Also, commonly used methods for estimating the
\(pMSE\), such as logistic regression and classification and regression
trees, are vulnerable to model misspecification
(\citeproc{ref-drechsler_utility_2022}{Drechsler 2022}). Lastly, the
output of global utility measures can be hard to interpret, and say
little about the regions in which the synthetic data do not resemble the
true data accurately enough.

To overcome some of the issues related to traditional global utility
measures, we approach global utility evaluation from the perspective of
density ratio estimation. The density ratio estimation framework
provides a theoretically and practically established toolbox for
comparing probability distributions without requiring separate
estimation of the underlying densities (for an overview, see
\citeproc{ref-sugiyama_suzuki_kanamori_2012}{Sugiyama, Suzuki, and
Kanamori 2012a}). Intuitively, if two data sets have similar
multivariate distributions, the density ratio should be close to one
over the range of the data. If the distributions of the observed and
synthetic data are very different, the density ratio should be far from
one at those regions where the distributions differ. Rather than
estimating the two densities separately and calculating their ratio,
density ratio estimation methods estimate this ratio directly in a
non-parametric way (e.g.,
\citeproc{ref-sugiyama_suzuki_kanamori_2012}{Sugiyama, Suzuki, and
Kanamori 2012a}; \citeproc{ref-sugiyama_kliep_2007}{Sugiyama et al.
2007}; \citeproc{ref-izbicki_dre_2014}{Izbicki, Lee, and Schafer 2014}).
This approach alleviates the challenging problem of estimating two
densities, which may magnify estimation errors when subsequently taking
their ratio. The density ratio framework is made readily available in
the \texttt{R}-package \texttt{densityratio}
(\citeproc{ref-densityratio}{Volker 2023}), which mitigates the issue of
model specification by providing automatic hyperparameter optimization.
Importantly, the density ratio is estimated over the entire range of the
data, which provides measures of utility at every (possible) point in
the data space. This point-specific quantification of utility turns out
to be a useful side-product, as it shows where the synthetic data
deviate and can thus guide improvements of the synthesis model.

In the remainder of the article, we first review the density ratio
framework. We discuss how divergences estimated through the density
ratio can be used as global utility measures, and relate these to
existing measures such as the \(pMSE\) and the Kullback-Leibler
divergence. Additionally, we show how the estimated density ratio itself
can be interpreted as a local utility measure. We empirically compare
density ratio-based measures with established utility measures in
simulations with continuous and mixed data, and apply density ratio
estimation in a case study where we evaluate the utility of multiple
synthetic versions of the U.S. Current Population Survey. We conclude
with a discussion of the results, highlight the strengths and weaknesses
of the density ratio framework, and provide recommendations for future
research.

\section{Background}\label{background}

Over the years, many methods have been introduced to generate synthetic
data, all with the aim of providing a suitable balance between privacy
and utility. These methods can be relatively simple, such as a sequence
of generalized linear models (e.g.,
\citeproc{ref-reiter_releasing_2004}{Reiter 2004}), or as complex as
deep learning models with thousands of parameters (e.g.,
\citeproc{ref-xu_ctgan_2019}{Xu et al. 2019}), with many options in
between. However, the complexity of the generation process is not
necessarily a good indicator of the quality of the synthetic data. That
is, relatively simple methods could still capture the most important
aspects of the data that complex methods fail to capture (and vice
versa; see, e.g., \citeproc{ref-foessing_evaluation_2024}{Fössing and
Drechsler 2024}). Hence, data providers typically do not know which
synthesis method will provide the highest utility a priori, and might
compare multiple synthesis strategies to determine which data set will
be released. Good global utility measures can help in this process, by
allowing to quantify the quality of the candidate synthetic data sets.
Moreover, such global utility measures may guide the synthesis process
itself, if they provide sufficiently specific information about the
degree of misfit of the synthetic data. Throughout, we focus on global
utility measures, since they can provide a general sense of the quality
of synthetic data sets. In many situations, the data provider does not
know which analyses will be performed with the synthetic data, and
analysis-specific utility measures thus lack generality. If the data
provider knows for which purposes the data will be used,
analysis-specific utility measures may be more informative and can thus
be used complementarily. In the upcoming section, we provide an overview
of existing global utility measures, and introduce the density ratio
framework as an encompassing approach to evaluating global utility.

\subsection{Existing global utility
measures}\label{existing-global-utility-measures}

Global utility measures typically attempt to quantify the distributional
similarity between the observed and synthetic data samples. The
intuition is that if two data sets are realizations from the same
underlying distribution, the data sets can be used for the same
purposes, and analyses on the two data sets should yield similar
results. One way to evaluate distributional similarity is by assessing
whether a classification model can tell samples from the two
distributions apart (see \citeproc{ref-kim_classification_2021}{Kim et
al. 2021}, who formalize the connection between classification accuracy
and two-sample testing). Hence, if a classification model can
distinguish between the observed and synthetic data with high accuracy,
the distributional similarity is low, and so is the global utility. If a
classifier cannot distinguish between the observed and synthetic data,
one would conclude that the global utility is high. The propensity score
mean-squared error (\(pMSE\)), introduced by Woo et al.
(\citeproc{ref-Woo_global_2009}{2009}) and further developed in Snoke et
al. (\citeproc{ref-snoke_utility_2018}{2018}), formalizes this
intuition. Let \(I_i\) denote an indicator variable that equals \(1\) if
observation \(i\) (\(i \in \{1, \dots, N\}\); \(N = \nobs + \nsyn\))
belongs to the synthetic data, and \(0\) otherwise. These indicators are
used in a classifier that outputs the predicted probability of
observation \(i\) being a synthetic record \(\hat{\pi}_i\) based on the
observation's scores on the variables (this can be the set of all
variables, but also a subset). From these, the utility statistic is
calculated as \begin{equation}\phantomsection\label{eq-pMSE}{
pMSE = \frac{1}{N} \sum_{i=1}^N \Big(\hat{\pi}_i - \frac{\nsyn}{N}\Big)^2,
}\end{equation} which ought to be smaller when the synthetic data is
more like the observed data. Crucially, the \(pMSE\) depends on the
classification model used and increases in the flexibility of the
classification model, making it prone to overfitting and hard to
interpret. To combat these issues, Snoke et al.
(\citeproc{ref-snoke_utility_2018}{2018}) suggest to compare the
\(pMSE\)-value with its expected value under the null hypothesis that
the observed and synthetic data are indistinguishable. Provided that the
classification model is a logistic regression model with \(k\)
parameters (including the intercept), Snoke et al.
(\citeproc{ref-snoke_utility_2018}{2018}) show that the expected
\(pMSE\) is given by \[
\mathbb{E}[pMSE] = \Big(\frac{k-1}{N}\Big) \Big(\frac{\nobs}{N}\Big)^2
\Big(\frac{\nsyn}{N}\Big).
\] For other classification models, the expectation can be approximated
through a resampling procedure. Accordingly, the \(pMSE\)-ratio is given
by \[
pMSE\text{-ratio} = \frac{pMSE}{\mathbb{E}[pMSE]}.
\] Apart from the \(pMSE\), several other measures can be constructed
from the estimated propensity scores, such as the percentage of records
correctly predicted (\citeproc{ref-raab2021assessing}{Raab, Nowok, and
Dibben 2021}) or the Kolmogorov-Smirnov statistic
(\citeproc{ref-Bowen_differentially_2021}{Bowen, Liu, and Su 2021}),
both of which are strongly correlated with the \(pMSE\)
(\citeproc{ref-raab2021assessing}{Raab, Nowok, and Dibben 2021}).

Due to its intuitive nature, multiple studies advise the use of the
\(pMSE\) as a promising technique to evaluate the quality of synthetic
data (e.g., \citeproc{ref-raab2017guidelines}{Raab, Nowok, and Dibben
2017}, \citeproc{ref-raab2021assessing}{2021};
\citeproc{ref-hu_advancing_2024}{Hu and Bowen 2024}). Yet, it is not
free of criticism. The usefulness of the \(pMSE\) hinges on choosing a
model that can capture the important intricacies of the observed data.
Drechsler (\citeproc{ref-drechsler_utility_2022}{2022}) illustrated that
the utility score is highly dependent on the model used to estimate the
propensity scores, and that clear improvements in synthesis models are
not necessarily picked up in the \(pMSE\). Moreover, selecting an
appropriate propensity score model may be difficult, as the challenges
associated with model selection, such as the bias-variance trade-off,
typically apply. Finally, the \(pMSE\) tends to favor synthesis models
that lie in the same model class as the model to construct the \(pMSE\)
(\citeproc{ref-drechsler_utility_2022}{Drechsler 2022}), which might
thus lead to an unjustified preference for similar synthesis models.

Another way to formalize distributional similarity is through the
Kullback-Leibler (KL) divergence, as proposed in Karr et al.
(\citeproc{ref-karr_utility_2006}{2006}). The KL-divergence measures the
relative entropy from the probability distribution of the observed data
\(\pobs(\bx)\) to the probability distribution of the synthetic data
\(\psyn(\bx)\) (with \(\bx \in \mathbb{R}^{D}\)), and is defined
as\footnote{In contrast to Karr et al.
  (\citeproc{ref-karr_utility_2006}{2006}), we define the KL-divergence
  as the relative entropy of \(\pobs\) with respect to \(\psyn\), such
  that it is consistent with our later formulations.}
\begin{equation}\phantomsection\label{eq-kl-div}{
KL(\pobs || \psyn) = 
\int \pobs(\bx) \log \frac{\pobs(\bx)}{\psyn(\bx)} \text{d} \bx.
}\end{equation} Karr et al. (\citeproc{ref-karr_utility_2006}{2006})
argue that the KL-divergence can be computed by approximating the
integral using density estimators \(\hat{p}_\text{obs}\) and
\(\hat{p}_\text{syn}\). An implementation of this approach for
divergence estimation has been described in Wang, Kulkarni, and Verdu
(\citeproc{ref-wang_divergence_2009}{2009}), where the densities are
estimated using nearest-neighbor density estimation. Another approach
posed by Karr et al. (\citeproc{ref-karr_utility_2006}{2006}) is to
assume multivariate normality for both \(\psyn\) and \(\pobs\), and
calculate the KL-divergence in closed form from the sample means and
covariance matrices. Although all these approaches are easy to
implement, they have their limitations. Density estimation is
challenging in high-dimensional settings, and assuming multivariate
normality might be too restrictive. We show in later sections that the
KL-divergence naturally fits into the density ratio estimation
framework.

\subsection{Density ratio estimation}\label{density-ratio-estimation}

\begin{figure}[t]

\centering{

\includegraphics[width=1\linewidth,height=\textheight,keepaspectratio]{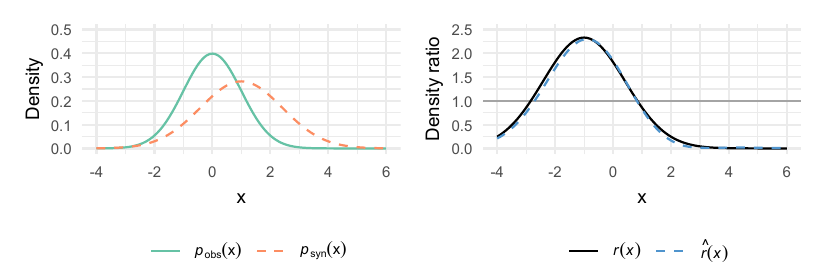}

}

\caption{\label{fig-dr-plot}Example of the true and estimated density
ratio of two normal distributions with different means and variances
(i.e., \(\pobs(\bx) = N(0,1)\) and \(\psyn(\bx) = N(1,2)\)). The
function \(r(\bx) = \pobs(\bx)/\psyn(\bx)\) denotes the true density
ratio, the function \(\hat{r}(\bx)\) denotes an estimate of the density
ratio based on \(\nsyn = \nobs = 200\) samples from each distribution
obtained with unconstrained Least-Squares Importance Fitting (uLSIF).
Note that the density ratio is itself not a density.}

\end{figure}

The density ratio estimation framework was originally developed in the
machine learning community for the comparison of two probability
distributions (for an overview, see
\citeproc{ref-sugiyama_suzuki_kanamori_2012}{Sugiyama, Suzuki, and
Kanamori 2012a}). The framework has been shown to be applicable to
prediction (\citeproc{ref-sugiyama_conditional_2010}{Sugiyama et al.
2010}; \citeproc{ref-sugiyama_classification_2010}{Sugiyama 2010}),
outlier detection (\citeproc{ref-shohei_dre_outlier_2008}{Hido et al.
2008}), change-point detection in time-series
(\citeproc{ref-liu_change_2013}{Liu et al. 2013}), importance weighting
under domain adaptation (i.e., sample selection bias;
\citeproc{ref-kanamori_ulsif_2009}{Kanamori, Hido, and Sugiyama 2009}),
and two-sample homogeneity tests
(\citeproc{ref-sugiyama_lstst_2011}{Sugiyama, Suzuki, et al. 2011}). The
general idea of density ratio estimation is depicted in
Figure~\ref{fig-dr-plot}, and boils down to comparing two distributions
by modelling the density ratio \(r(\bx)\). In our case, we define
\(r(\bx)\) as the ratio between the probability distributions of the
numerator samples, taken from the observed data distribution,
\(\pobs(\bx)\), and the denominator samples, taken from the synthetic
data distribution, \(\psyn(\bx)\), assumed to encompass the support of
the observed data, such that
\begin{equation}\phantomsection\label{eq-dr}{
r(\bx) = \frac{\pobs(\bx)}{\psyn(\bx)}.
}\end{equation} This specification has the intuitive interpretation that
in locations where the density ratio takes values larger than 1, too few
synthetic observations are generated in that region. Conversely, in
locations where the density ratio takes values smaller than one, too
many synthetic observations are generated in that region (see
Figure~\ref{fig-dr-plot}). We take the current form of the density
ratio, instead of its reciprocal, for two reasons. First, this
formulation corresponds to the familiar form of the Kullback-Leibler
divergence, where the approximating distribution is placed in the
denominator, whereas the numerator distribution is the reference
distribution. Second, the synthetic data commonly has broader support
than the observed data, for example because the observed data have a
bounded domain while the synthetic data are drawn from an unbounded
probability distribution.

A straightforward approach to estimating \(r(\bx)\) from samples of
\(\pobs(\bx)\) and \(\psyn(\bx)\) would be to estimate the observed and
synthetic data density separately, for example using kernel density
estimation (e.g., see \citeproc{ref-Scott1992}{Scott 1992} for an
overview), and subsequently compute the ratio of these estimated
densities. However, density estimation is one of the hardest tasks in
statistical learning, unavoidably leading to estimation errors for both
densities, especially in high dimensions
(\citeproc{ref-sugiyama_suzuki_kanamori_2012}{Sugiyama, Suzuki, and
Kanamori 2012a}). When subsequently taking the ratio of the estimated
densities, estimation errors tend to be magnified. Direct density ratio
estimation avoids this issue by specifying and estimating a model
directly for the ratio without first estimating the separate densities.
Extensive simulations on a wide variety of tasks showed that this
approach typically outperforms density ratio estimation through kernel
density estimation for each distribution separately, especially when the
dimensionality of the data increases (e.g.,
\citeproc{ref-Kanamori2012}{Kanamori, Suzuki, and Sugiyama 2012b};
\citeproc{ref-shohei_dre_outlier_2008}{Hido et al. 2008};
\citeproc{ref-kanamori_ulsif_2009}{Kanamori, Hido, and Sugiyama 2009}).

\linespread{1}

\linespread{2}

\subsubsection{Estimating the density
ratio}\label{estimating-the-density-ratio}

Over the past years, several methods for direct density ratio estimation
have been developed. A large class of these methods attempts to directly
minimize the error between the true density ratio \(r(\bx)\) and a
density ratio model \(\hat{r}(\bx)\). Following this approach, we define
a loss function \(\mathcal{L}(r(\bx), \hat{r}(\bx))\) that measures the
discrepancy between the true and estimated density ratio. To give an
example, consider the following loss based on the squared error
\begin{equation}\phantomsection\label{eq-squared-error}{
\begin{aligned}
\mathcal{L}^*_{S}(r(\bx), \hat{r}(\bx)) &= 
\frac{1}{2} \int (\hat{r}(\bx) - r(\bx))^2 \psyn(\bx) \text{d}\bx \\
&= \frac{1}{2} \int \hat{r}(\bx)^2 \psyn(\bx) \text{d}\bx - 
\int \hat{r}(\bx) r(\bx) \psyn(\bx) \text{d}\bx + 
\frac{1}{2} \int r(\bx)^2 \psyn(\bx) \text{d}\bx.
\end{aligned}
}\end{equation} The second term can be rewritten, because the
denominator in \(r(\bx)\) cancels with \(\psyn(\bx)\), while the third
term is a constant with respect to the parameters in the density ratio
model and can thus be ignored. Hence, we are left with the following
objective to minimize
\begin{equation}\phantomsection\label{eq-squared-error-loss}{
\mathcal{L}_{S} (r(\bx), \hat{r}(\bx)) = \frac{1}{2} \int \hat{r}(\bx)^2 \psyn(\bx) \text{d}\bx - \int \hat{r}(\bx) \pobs(\bx) \text{d}\bx,
}\end{equation} which is termed least squares importance fitting (LSIF)
by Kanamori, Hido, and Sugiyama
(\citeproc{ref-kanamori_ulsif_2009}{2009}). When the parameters of the
density ratio are not constrained to be positive, this loss yields a
closed-form solution to the density ratio, which is commonly called
unconstrained LSIF (uLSIF), which we will use throughout.

This approach can be straightforwardly generalized to a wide class of
loss functions that fall under the family of Bregman divergences (see
\citeproc{ref-sugiyama_bregman_2012}{Sugiyama, Suzuki, and Kanamori
2012b}; \citeproc{ref-mohamed2017learning}{Mohamed and Lakshminarayanan
2017}). Then, a general class of losses is encompassed by the expression
\begin{equation}\phantomsection\label{eq-bregman-loss}{
\mathcal{L}^*_f (r(\bx), \hat{r}(\bx)) =
\int \Big(
f(r(\bx)) - f(\hat{r}(\bx)) - f'(\hat{r}(\bx))(r(\bx) - \hat{r}(\bx))
\Big) \psyn(\bx) \text{d}\bx,
}\end{equation} where \(f\) is a differentiable and strictly convex
function with derivative \(f'\). Then, ignoring the terms independent of
\(\hat{r}(\bx)\) and noting that \(r(\bx)\psyn(\bx) = \pobs(\bx)\), we
obtain the following objective
\begin{equation}\phantomsection\label{eq-bregman-objective}{
\mathcal{L}_f (r(\bx), \hat{r}(\bx)) = \int 
\Big( 
f'(\hat{r}(\bx)) \hat{r}(\bx) - f(\hat{r}(\bx))
\Big)  \psyn(\bx) \text{d}\bx
- \int f'(\hat{r}(\bx)) \pobs(\bx)
\text{d}\bx.
}\end{equation} Minimizing this loss over \(\hat{r}(\bx)\) for different
functions \(f\) yields different estimators for the density ratio, that
focus on different regions of the density ratio.\footnote{It is easy to
  see that using \(f(x) = \frac{1}{2}(x-1)^2\) turns the Bregman
  divergence (Equation~\ref{eq-bregman-loss}) into the squared error
  (Equation~\ref{eq-squared-error}).} Specifically, some estimators
place more emphasis on accurately modelling the regions in which the
true density ratio is large, but less emphasis on accurately estimating
the density ratio in regions where it takes small values, and vice versa
(see \citeproc{ref-sugiyama_bregman_2012}{Sugiyama, Suzuki, and Kanamori
2012b}; \citeproc{ref-menon2016dreloss}{Menon and Ong 2016}).
Importantly, Sugiyama, Suzuki, and Kanamori
(\citeproc{ref-sugiyama_bregman_2012}{2012b}) show that the Bregman
divergence minimization approach to density ratio estimation is
equivalent to estimating an \(f\)-divergence
(\citeproc{ref-ali_silvey_divergence_1966}{Ali and Silvey 1966}) of the
observed data distribution from the synthetic data distribution. Hence,
when estimating the density ratio, one implicitly estimates the
divergence between the observed and synthetic data distributions. When
using \(f(x) = \frac{1}{2} (x-1)^2\), this divergence is the Pearson
divergence.

After defining a loss function, we need a model for the density ratio
function \(\hat{r}(\bx)\). There are many possibilities to specify this
model, but a common choice is to use a linear model for the density
ratio (e.g., \citeproc{ref-huang_kmm_2006}{Huang et al. 2006};
\citeproc{ref-kanamori_ulsif_2009}{Kanamori, Hido, and Sugiyama 2009};
\citeproc{ref-izbicki_dre_2014}{Izbicki, Lee, and Schafer 2014};
\citeproc{ref-gruber2024overcoming}{Gruber et al. 2024}). That is, we
define the density ratio model as
\begin{equation}\phantomsection\label{eq-dr-model}{
\hat{r}(\bx) = \boldsymbol{\varphi}(\bx) \boldsymbol{\theta},
}\end{equation} where \(\boldsymbol{\varphi}(\bx)\) is a basis function
vector (optionally including an intercept term), that transforms the
data from a \(D\)-dimensional to a \(b\)-dimensional space, and
\(\boldsymbol{\theta}\) is a \(b\)-dimensional parameter vector.
Although the model is linear in its parameters, the density ratio itself
is typically a non-linear function of the data due to the basis
functions. The parameter vector \(\boldsymbol{\theta}\) is estimated to
minimize the discrepancy with the true density ratio using the loss
function \(\mathcal{L}(r(\bx), \hat{r}(\bx))\) in
Equation~\ref{eq-bregman-objective}. Potentially, the loss function can
be extended with a regularization term (e.g., an \(L_1\) or \(L_2\) norm
on the parameters of the model), to decrease the flexibility of the
density ratio.

Also for the basis functions, several specifications are possible,
ranging from an identity function
(\citeproc{ref-qin_inferences_1998}{Qin 1998}) to normalizing flows
(\citeproc{ref-choi_featurized_2021}{Choi, Liao, and Ermon 2021}) or
neural networks (\citeproc{ref-tiao2018dre}{Tiao 2018};
\citeproc{ref-uehara2016generative}{Uehara et al. 2016}). However, we
limit our discussion and simulation to the Gaussian kernel function,
which is a commonly used basis function in the density ratio literature
(e.g., \citeproc{ref-huang_kmm_2006}{Huang et al. 2006};
\citeproc{ref-sugiyama_kliep_2007}{Sugiyama et al. 2007};
\citeproc{ref-kanamori_ulsif_2009}{Kanamori, Hido, and Sugiyama 2009};
\citeproc{ref-liu_change_2013}{Liu et al. 2013};
\citeproc{ref-gruber2024overcoming}{Gruber et al. 2024}). The Gaussian
kernel function is defined as \[
\boldsymbol{\varphi}_j(\bx) = \mathcal{K}(\bx, \boldsymbol{c}_j) = 
\exp\left(
- \frac{\lVert\bx - \boldsymbol{c}_j \rVert^2}{2 \sigma^2}
\right),
\] where \(\boldsymbol{c}_j\) (\(j \in 1, \dots, b\)) denotes the
centers of the Gaussian kernel functions, and \(\sigma\) controls the
kernel width (i.e., it defines over which distance similarity between
the observations and the centers are considered relevant; for more
information on kernel functions, see, e.g.,
\citeproc{ref-murphy_pmlintro_2022}{Murphy 2022}). Due to this choice of
kernel, the density ratio is a smooth function, and differences in
support between the observed and synthetic data typically have limited
impact. That is, while the true density ratio is unbounded when the
observed data have mass where the synthetic data have none, the
estimated density ratio remains finite because the density ratio is
shrunk towards zero. To improve estimation accuracy and stability, it is
possible to use relative estimators of the density ratio
(\citeproc{ref-yamada_rulsif_2013}{Yamada et al. 2013}), which ensure
that the denominator is non-zero when the numerator is or alternative
density ratio estimation techniques (e.g.,
\citeproc{ref-rhodes_telescoping_2020}{Rhodes, Xu, and Gutmann 2020}).
The appropriate width of the kernel and the optimal value of the
regularization parameter can be determined using cross-validation. The
centers are typically sampled from the data (in our case, both the
observed and synthetic data).

After defining the model and the loss function, the density ratio can be
estimated. The \texttt{densityratio} package in R
(\citeproc{ref-densityratio}{Volker 2023}) provides an implementation of
commonly used loss functions with a Gaussian kernel basis function. The
package comes with an easy-to-use interface, automatic cross-validation
of hyperparameters and builds on \texttt{C++} for fast computation.
Alternatively, parameter estimation can be carried out using standard
optimization techniques (see Appendix \ref{sec-app-A} for a brief
comparison of both approaches). The density ratio estimation framework
thus equips researchers with a measure of utility which is accurate,
fast to compute and does not require user specification of the model.

\subsubsection{Evaluating data utility with the density
ratio}\label{evaluating-data-utility-with-the-density-ratio}

Having estimated the density ratio, we can evaluate both global and
local utility of the synthetic data. Like other global utility measures,
the density ratio can be used to construct a single discrepancy measure
that quantifies the degree of misfit of the synthetic data. The density
ratio can be used to calculate an empirical \(f\)-divergence between the
observed and synthetic data distributions, which serves as a measure of
utility for the synthetic data. For example, the empirical Pearson
divergence follows directly from the squared error in
Equation~\ref{eq-squared-error} \[
\widehat{PE} = \frac{1}{\nobs} \sum_{i=1}^{\nobs} \hat r(\bx_{\text{obs}_i}) - \frac{1}{2\nsyn} \sum_{i'=1}^{\nsyn} \hat r(\bx_{\text{syn}_{i'}})^2 - 1/2.
\] Analogous expressions exist for other members of the family of
\(f\)-divergences. Although such divergence statistics are difficult to
interpret in an absolute sense, they can be used as a relative measure
of the quality of the synthetic data. That is, for different synthetic
data sets, we can calculate the divergence to the observed data, and
compare these values to determine which synthetic data set is most
similar to the observed data. More formally, \(f\)-divergences can be
used in a permutation test to determine whether the synthetic data is
generated from the same distribution as the observed data. The
corresponding \(p\)-value can be calculated by comparing the observed
divergence measure to the null distribution of divergence measures
obtained using random permutations of the observed and synthetic data
(see, for example, \citeproc{ref-sugiyama_lstst_2011}{Sugiyama, Suzuki,
et al. 2011}; \citeproc{ref-Wornowizki2016}{Wornowizki and Fried 2016};
\citeproc{ref-kanamori_divergence_2012}{Kanamori, Suzuki, and Sugiyama
2012a}).

An appealing characteristic of the density ratio framework is that it
also readily allows for evaluating local utility. As said, the estimated
density ratio shows in which regions the synthetic data diverges from
the observed data. By inspecting which observations have a high, or
small, estimated density ratio, we can identify the regions in which the
synthetic data is most different from the observed data. Visual
inspection can aid this process, for example by plotting the estimated
density ratio against variables or pairs of variables in the data. On an
even finer scale, the density ratio can be used to identify individual
synthetic records that are far from what can be expected on the basis of
the observed data. This information could identify flaws in the
synthetic data model, or could be used directly to remove such
observations from the synthetic data. We illustrate these features in
Section~\ref{sec-app}.

A final advantage of the density ratio estimation framework lies in its
flexibility. The combination of a non-parametric kernel-based model with
automatic cross-validation for hyperparameter selection (e.g., kernel
and regularization parameters) allows for flexible estimation of the
density ratio. This flexibility reduces the burden on the user in terms
of model specification when evaluating the utility of the synthetic
data. All functionality described above, including visualization,
divergence estimation (and testing), extracting importance weights and
automatic cross-validation for hyperparameter selection, is readily
available in the \texttt{densityratio} \texttt{R}-package
(\citeproc{ref-densityratio}{Volker 2023}). The package provides fast
and efficient algorithms for density ratio estimation, and uses default
implementations that work across a wide range of data distributions,
which will be shown in the upcoming simulations.

\section{Numerical illustrations: Simulation
studies}\label{numerical-illustrations-simulation-studies}

To evaluate the performance of density ratio estimation and compare it
to existing methods, we conduct a series of simulation studies (all code
is available on GitHub\footnote{\url{https://github.com/thomvolker/dr-utility}}).
First, we illustrate the approach in a univariate simulation, after
which we evaluate the performance of the approach in a multivariate
setting, varying the distributions, number of variables and sample
sizes. To perform density ratio estimation, we use unconstrained
Least-Squares Importance Fitting (uLSIF;
\citeproc{ref-kanamori_ulsif_2009}{Kanamori, Hido, and Sugiyama 2009})
as implemented in the \texttt{densityratio} package in \texttt{R}
(\citeproc{ref-densityratio}{Volker 2023}), because it is fast and has
been shown to perform well in a variety of settings (e.g.,
\citeproc{ref-kanamori_ulsif_2009}{Kanamori, Hido, and Sugiyama 2009};
\citeproc{ref-li_application_2010}{Li et al. 2010};
\citeproc{ref-Kanamori2012}{Kanamori, Suzuki, and Sugiyama 2012b}). We
compare the performance of uLSIF to the \(pMSE\) (implemented in
\texttt{synthpop}; \citeproc{ref-synthpop}{Nowok, Raab, and Dibben
2016}), Kullback-Leibler divergence estimation through \(k\)-nearest
neighbor density estimation (implemented in \texttt{kldest};
\citeproc{ref-kldest}{Hartung 2024}) and the maximum mean discrepancy
(\(MMD\), as implemented in \texttt{kernlab};
\citeproc{ref-kernlab}{Karatzoglou et al. 2004};
\citeproc{ref-gretton_kernel_2012}{Gretton et al. 2012}). Note that
Kullback-Leibler divergence estimation through \(k\)-nearest neighbor
density estimation implicitly performs density ratio estimation (see
Appendix \ref{sec-app-B}). We additionally evaluate how sensitive the
density ratio is to model specification details in Appendix
\ref{sec-app-stability}. Each simulation is repeated 500 times.

\subsection{Univariate illustrations}\label{sec-univ-sims}

\begin{figure}[t]

\centering{

\includegraphics[width=1\linewidth,height=\textheight,keepaspectratio]{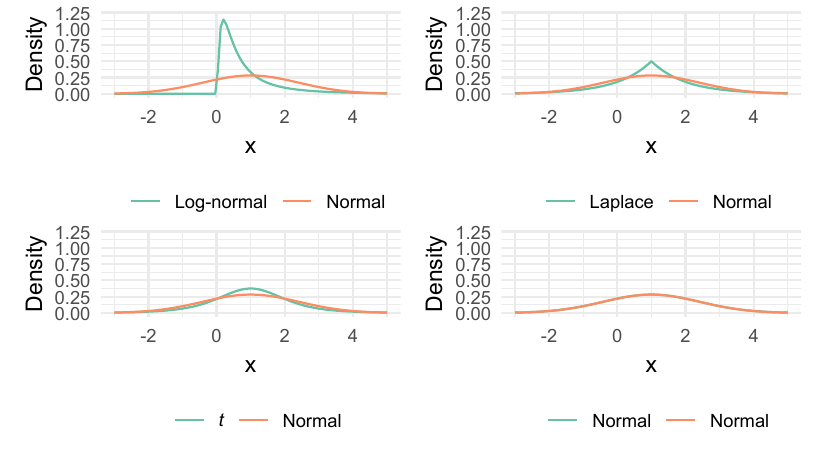}

}

\caption{\label{fig-densities-sim1}True and synthetic data densities for
the four simulations with Log-normal, Laplace, location-scale \(t\)- and
Normal densities. All data-generating mechanisms have the same mean
\(\mu = 1\) and variance \(\sigma^2 = 2\). Note that the true and
synthetic data density in the bottom right panel are completely
overlapping.}

\end{figure}

We start with simulations in a univariate setting, which allows us to
build some intuition by visualizing the estimated density ratios. We
consider four settings, where \(\nobs = 250\) real samples are generated
from (1) a log-normal distribution, (2) a Laplace distribution, (3) a
location-scale \(t\)-distribution and (4) a normal distribution (see
Figure~\ref{fig-densities-sim1}), with the following parameters:

\begin{enumerate}
\def\labelenumi{\arabic{enumi}.}
\tightlist
\item
  Log-normal distribution with log-mean
  \(\mu_{\text{log}} = \log \{1/\sqrt{3} \}\) and log-variance
  \(\sigma^2_\text{log} = \log 3\).
\item
  Laplace distribution with location parameter \(\mu = 1\) and scale
  \(b = 1\).
\item
  Location-scale \(t\)-distribution with location parameter \(\mu = 1\),
  scale \(\tau^2 = 1\) and degrees of freedom \(\nu = 4\).
\item
  Normal distribution with mean \(\mu = 1\) and variance
  \(\sigma^2 = 2\).
\end{enumerate}

The synthetic data is generated from an approximating Gaussian
distribution with its mean and variance estimated from the observed
data. Hence, while the synthetic data has the same mean and variance as
the true data, higher-order moments differ, except when the true data is
also normally distributed. These settings illustrate commonly
encountered scenarios in the synthetic data field, in the sense that the
true data generating mechanism is unknown and needs to be approximated
using a simpler approximating model. The level of misspecification can
range from quite severe to virtually absent, typically depending on the
complexity of the data. For example, bounds or logical constraints in
the real data can be accidentally omitted in the synthesis model,
leading to relatively large discrepancies in some regions of the
distributions. The density ratio model is estimated using uLSIF with the
default Gaussian kernel and \(L_2\)-penalty on the parameter vector
\(\boldsymbol{\theta}\). Both the kernel bandwidth and the
regularization parameter are selected using cross-validation, using the
default settings in the \texttt{densityratio} package.

\linespread{1}

\linespread{2}

\linespread{1}

\begin{figure}[t]

\centering{

\includegraphics[width=1\linewidth,height=\textheight,keepaspectratio]{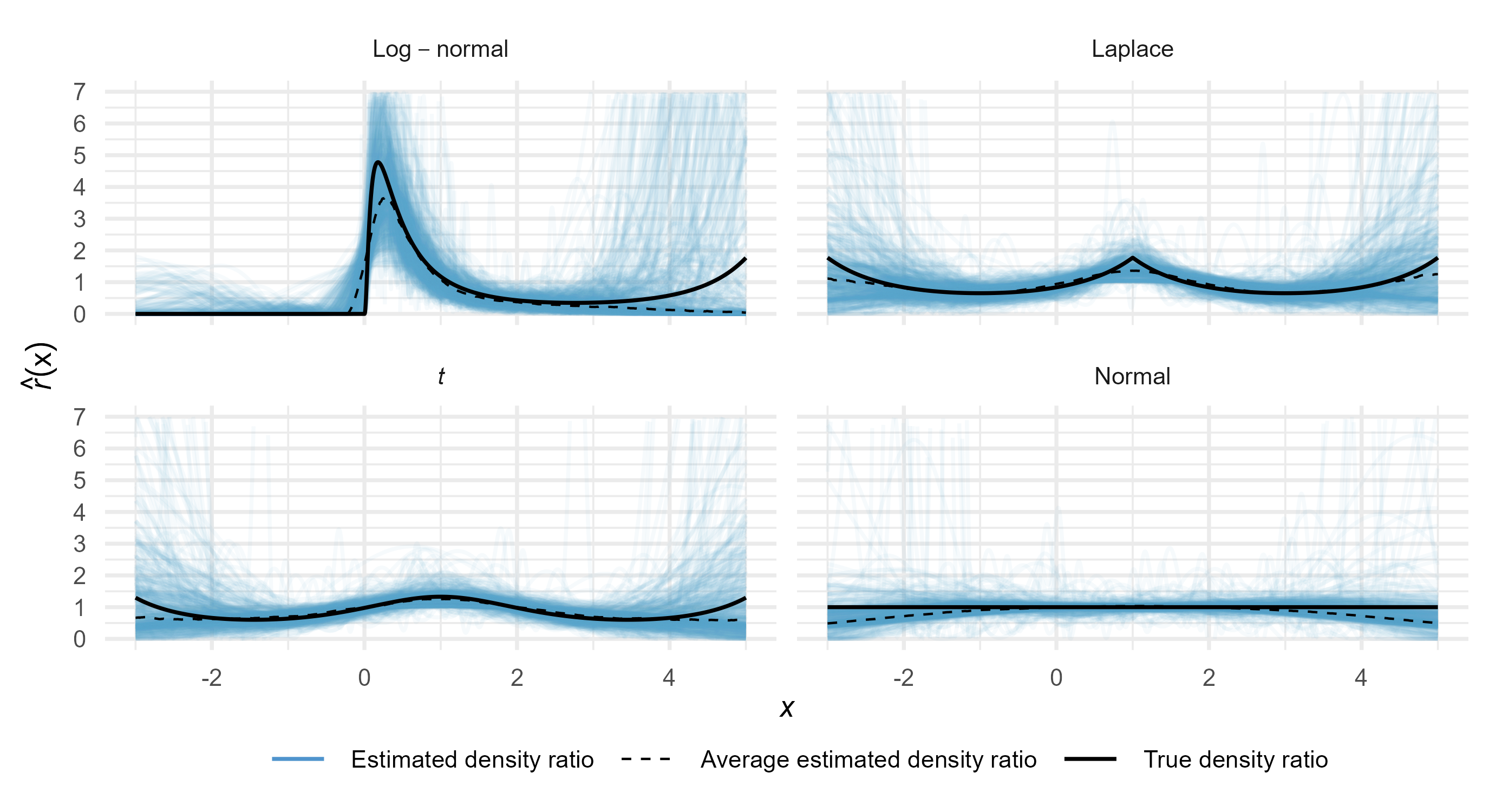}

}

\caption{\label{fig-sim1-results}Estimated density ratios by
unconstrained least-squares importance fitting in four univariate
examples: A log-normal distribution, a Laplace distribution, a
location-scale \(t\)-distribution and a normal distribution, all
approximated by a normal distribution with the same mean and variance as
the sample from the true distribution. Note that the mass of the
synthetic data distribution in the tails (smaller than \(-2\) or greater
than \(4\)) is smaller than \(0.034\).}

\end{figure}%

\linespread{2}

Figure~\ref{fig-sim1-results} shows that the estimated density ratios
(the blue lines) follow the general trend of the true density ratio (the
solid black line) for each data-generating mechanism. Especially in the
center of each panel, approximately between \(-2\) and \(4\), the
average density ratio (the dashed black line) closely matches the true
density ratio. On average, all estimated density ratios model the peak
in the center of each panel, where the true data generating distribution
has more mass than the synthetic data distribution (except the bottom
right panel, where the synthetic data distribution is correct and the
density ratio equals one). When moving from the middle to the sides,
where the synthetic data distributions have more mass, the estimated
density ratios still follow the true density ratios, in the sense that
they are typically smaller than \(1\). In the tails of the distribution,
where the synthetic data distribution has almost no mass
(\(P(x \leq -2) = P(x \geq 4) \approx 0.017\)), estimating the density
ratio becomes difficult, leading to estimated density ratios that
diverge. In these regions, the density ratio is underestimated on
average due to the regularization, but with substantial variability.
Finally, the panels show that sharp transitions, such as the cut-off at
zero for the log-normal distribution and the peak of the Laplace
distribution are smoothed out to some extent.

We briefly compare the performance of uLSIF to \(k\)-nearest neighbor
density ratio estimation, which is implicitly used in Kullback-Leibler
divergence estimation through \(k\)-nearest neighbor density estimation
(see Appendix \ref{sec-app-B} for details). We set the number of
neighbors to \(k = 15 \approx \sqrt n\) (as suggested by
\citeproc{ref-loftsgaarden_nonparametric_1965}{Loftsgaarden and
Quesenberry 1965}). Figure~\ref{fig-AppB} shows that the estimated
density ratios by uLSIF are substantially less variable in the center of
the distribution than the estimated density ratios by \(k\)-nearest
neighbor density ratio estimation, whereas uLSIF is more variable in the
tails. The estimated density ratios by \(k\)-nearest neighbor density
ratio estimation have substantial shrinkage in the tails of the
distribution, where data is scarce. We conclude that uLSIF is able to
provide more accurate estimates of the density ratio if there are enough
samples in the regions of interest.

\subsection{Stability of density ratio estimation for different model
specifications}\label{stability-of-density-ratio-estimation-for-different-model-specifications}

Highly accurate estimates of the density ratio may be out of reach when
data are scarce. However, for utility evaluation, it typically suffices
to obtain an accurate summary statistic of the discrepancy between the
observed and synthetic data, such as a divergence measure. This
statistic should not depend too strongly on details of the model
specification that are hard to justify. To evaluate the sensitivity of
density ratio-based utility measures to model specification, we generate
five synthetic data sets for each observed data set from a mixture of
the true and approximating normal data distributions with size
\(\nobs = \nsyn \in \{250, 500\}\). The synthetic data distribution is
defined as \((1-p)\) times the true data distribution plus \(p\) times
the approximating Gaussian distribution, for
\(p \in \{0, 0.25, 0.5, 0.75, 1\}\), thus ranging from the true
distribution (\(p=0\)) to the approximating normal distribution
(\(p=1\)). The divergence between the synthetic and observed data is
estimated under various density ratio model specifications: default
uLSIF with automatic hyperparameter tuning, fewer kernel centers, fixed
hyperparameters yielding a more flexible specification, relative uLSIF
(\citeproc{ref-yamada_rulsif_2013}{Yamada et al. 2013}), KLIEP, which
uses a KL-loss to estimate the density ratio
(\citeproc{ref-sugiyama_kliep_2007}{Sugiyama et al. 2007}), and
KL-estimation using \(k\)-nearest neighbors density ratio estimation.
Within each simulation run, we compute the rank correlations between the
divergences obtained with the different specifications and with the
mixing proportion \(p\), and compare with various traditional utility
measures (namely the \(pMSE\), and the \(MMD\) as implemented in
\texttt{kernlab}; \citeproc{ref-kernlab}{Karatzoglou et al. 2004};
\citeproc{ref-gretton_kernel_2012}{Gretton et al. 2012}). Details about
this simulation and the results are provided in Appendix
\ref{sec-app-stability}.

The divergence increases with the mixing proportion \(p\) on average in
all scenarios except the normal one, where each synthetic data model is
correctly specified. The uLSIF-specifications correlate strongly with
each other and with the other utility measures, especially for larger
sample sizes, except when the hyperparameters are fixed. For the latter,
the correlation with the mixing proportion is also lower. KLIEP
correlates less with uLSIF and with the mixing proportion. Even in the
normal scenario, all methods are moderately to strongly correlated,
indicating that they pick up the same deviations to some extent.

\subsection{Multivariate simulations}\label{multivariate-simulations}

We further investigate the performance of density ratio estimation in
multivariate simulations and compare its performance with existing
global utility measures. In these simulations, we evaluate whether
density ratio estimation captures improvements in synthetic data models
and reflects these improvements in the utility statistics. In what
follows, we describe the data-generating mechanisms, the synthetic data
models, the utility measures used to evaluate the quality of the
synthetic data models and the results of the simulations.

\subsubsection{True data-generating
mechanism}\label{true-data-generating-mechanism}

In the first set of multivariate simulations, we generate the real data
according to three mechanisms, consisting of \(D \in \{5, 25, 50\}\)
variables. When \(D = 5\), the first four variables are multivariate
normally distributed. These normally distributed variables have mean
\(\mu_d = 0\), variance \(\sigma^2_{d,d} = 1\) and covariance
\(\sigma^2_{d,d'} = 0.5\), for \(d = 1, \dots, D-1\). The fifth variable
is non-linearly related to the first variable, such that
\(X_5 \sim \mathcal{N}(X_1^2, \hat{\text{Var}}(X_1^2))\). When
\(D = 25\) or \(D = 50\), the first \(20\) and \(45\) variables are
distributed normally, again with zero mean, unit variance and covariance
\(\sigma_{d,d'} = 0.5\) for \(d = 1, \dots, D-5\). The last five
variables have a non-linear relationship with the first five variables,
defined as
\(X_{D-5+j} \sim \mathcal{N}(X_j^{j+1}, \hat{\text{Var}}(X_j^{j+1}))\),
for \(j = 1, \dots, 5\). Hence, the first variable with a non-linear
relationship is obtained by raising the first variable to the power
\(2\), the second is obtained by raising the second variable to the
power \(3\), and so on, with an additional variance term that scales
with the variability due to raising terms to a higher power. For each
number of variables, we consider sample sizes
\(\nobs = \nsyn \in \{100, 1000, 2000\}\).

In the second set of multivariate simulations, we simulate mixed data,
where the first three variables are categorical variables with
\(R_1 = 2\), \(R_2 = 4\) and \(R_3 = 12\) distinct levels, rendering 96
combinations in total. The second and third categorical variable depend
on the first categorical variable. Subsequently, we generate five
continuous variables conditional on the observed category, with means
depending on the third categorical variable, and different
covariance-structures depending on the first categorical variable (see
Appendix \ref{sec-app-multivariate} for details). We again use
\(\nobs = \nsyn \in \{100, 1000, 2000\}\).

\subsubsection{Synthetic data
generation}\label{synthetic-data-generation}

We generate synthetic data via three mechanisms, each incrementally
closer to the true data generating mechanism. In the continuous setting,
the first synthetic data model is a multivariate normal distribution
with means and variances estimated from the real data, but neither
covariances nor non-linear relationships are taken into account. The
second synthetic data model is a multivariate normal distribution with
means, variances and covariances estimated from the real data, but
without the non-linear relationships. The final, and correctly
specified, synthetic data model estimates the means, variances,
covariances and non-linear parameters from the data.

For the mixed setting, we likewise consider three incrementally
improving synthesis methods. In the first setting, the block of
continuous variables is correctly modelled, but the block of categorical
variables is generated independently of each other and of the continuous
variables. In the second setting, the block of continuous variables is
again modelled correctly, and the block of categorical variables is
jointly generated correctly, but jointly independent of the continuous
variables. The third synthesis strategy again samples from a correctly
specified model.

\subsubsection{Utility measures}\label{utility-measures}

The goal of the simulation study is to evaluate whether the global
utility measures are able to detect the improvements in the synthesis
models. We focus on four different utility measures: the Pearson
divergence, the KL-divergence estimated using \(k\)-nearest neighbors,
the \(pMSE\) and the \(MMD\). First, we use uLSIF with automatic
hyperparameter selection through cross-validation (the default in the
\texttt{densityratio}-package) and a fixed, flexible specification
without hyperparameter tuning. All details and additional results on the
robustness to model specification can be found in Appendix
\ref{sec-app-multivariate}. We express the estimated utility of the
synthetic data through the Pearson divergence (\(PE\);
\citeproc{ref-yamada_rulsif_2013}{Yamada et al. 2013}). We again compare
against the KL-divergence based on \(k\)-nearest neighbor density ratio
estimation, using \(k = 1\) (the default in the software) and
\(k = \sqrt{n}\) (as advocated by
\citeproc{ref-loftsgaarden_nonparametric_1965}{Loftsgaarden and
Quesenberry 1965}). Additionally, we use the \(pMSE\)-ratio, based on
the default CART model in \texttt{synthpop}, a CART model with smaller
complexity parameter (using \(cp = 10^{-6}\)) and a logistic regression
model with and without two-way interactions. Interactions are included
when the number of parameters is smaller than \(\nobs\). Finally, we
compare the results with the maximum mean discrepancy
(\citeproc{ref-kernlab}{Karatzoglou et al. 2004}). Categorical variables
are included as dummy variables and continuous variables in all
distance-based measures.

\subsubsection{Results 1: Continuous
data}\label{results-1-continuous-data}

\linespread{1}

\begin{table}

\caption{\label{tbl-sim2-results}Proportion of simulations (Setting 1)
in which the true synthetic data model order is captured by the utility
measure, for default uLSIF (\(PE\)), with fixed hyperparameters
(\(PE_\text{fixed}\)), KL-divergence based on its nearest neighbor
(\(KL_1\)) and based on its \(\sqrt{n}\)th nearest neighbor, maximum
mean discrepancy (\(MMD\)) and \(pMSE\)-ratio based on a CART model with
small complexity parameter (\(pMSE_{\text{CART}_{cp}}\)), based on a
logistic model with main effects (\(pMSE_\text{logit}\)) and a logit
model with main and interaction effects (\(pMSE_\text{logit, int}\)),
which is only estimated when the number of parameters is smaller than
\(\nobs\).\vspace{0.4cm}}

\centering{

\begin{tabularx}{\linewidth}{>{\raggedright\arraybackslash}X>{\raggedright\arraybackslash}X>{\centering\arraybackslash}p{10mm}>{\centering\arraybackslash}p{10mm}>{\centering\arraybackslash}p{10mm}>{\centering\arraybackslash}p{10mm}>{\centering\arraybackslash}p{10mm}>{\centering\arraybackslash}p{17mm}>{\centering\arraybackslash}p{17mm}>{\centering\arraybackslash}p{17mm}}
\toprule
$n$ & $D$ & $PE$ & $PE_{\text{fixed}}$ & $KL_1$ & $KL_{\sqrt{n}}$ & $MMD$ & $pMSE_{\text{CART}_{cp}}$ & $pMSE_{\text{logit}}$ & $pMSE_{\text{logit, int}}$\\
\midrule
100 & 5 & 0.65 & 0.84 & 0.86 & 0.86 & 0.75 & 0.35 & 0.22 & 0.36\\
100 & 25 & 0.63 & 0.68 & 0.83 & 0.93 & 0.63 & 0.32 & 0.23 & -\\
100 & 50 & 0.54 & 0.59 & 0.68 & 0.83 & 0.51 & 0.37 & 0.34 & -\\
\addlinespace
1000 & 5 & 1.00 & 1.00 & 1.00 & 1.00 & 1.00 & 0.95 & 0.21 & 0.31\\
1000 & 25 & 1.00 & 1.00 & 1.00 & 1.00 & 1.00 & 0.92 & 0.24 & 0.02\\
1000 & 50 & 0.98 & 1.00 & 1.00 & 1.00 & 0.94 & 0.87 & 0.33 & -\\
\addlinespace
2000 & 5 & 1.00 & 1.00 & 1.00 & 1.00 & 1.00 & 0.97 & 0.16 & 0.30\\
2000 & 25 & 1.00 & 1.00 & 1.00 & 1.00 & 1.00 & 0.99 & 0.25 & 0.01\\
2000 & 50 & 1.00 & 1.00 & 1.00 & 1.00 & 1.00 & 0.96 & 0.31 & 0.01\\
\bottomrule
\end{tabularx}

}

\end{table}%

\linespread{2}

Table~\ref{tbl-sim2-results} shows the proportion of simulations in
which the utility measures correctly order the synthetic data sets in
terms of quality of the synthesis model. That is, the first synthesis
model, using only the means and variances of each variable, should have
the lowest utility (highest divergence, \(pMSE\)-ratio or \(MMD\)),
followed by the second synthesis model, using the correct means,
variances and covariances but not the non-linear relationships, and the
third, correctly specified synthesis model should obtain the lowest
discrepancy. Table~\ref{tbl-sim2-results} shows that the KL-divergence
based on \(k = \sqrt{n}\) neighbors correctly ranks the synthetic data
sets most often, followed by the KL-divergence based on \(k = 1\)
neighbors, the uLSIF models and the \(MMD\). Unless the sample size is
small, these approaches correctly rank the synthetic data sets in almost
all instances. The performance of uLSIF is also relatively robust to
small differences in specification (see also Appendix
\ref{sec-app-multivariate}, Table~\ref{tbl-app-multivariate}), although
a more flexible specification yields somewhat better results when the
sample size is small, while KLIEP performs substantially worse than
uLSIF, except for the smallest sample size and the smallest number of
variables.

Essentially all methods outperform the \(pMSE\). For the logistic
models, this is to be expected, as main effects (and two-way
interactions) cannot distinguish between data sets with correct means
(and covariances). In fact, when means and covariances are modelled
correctly, also modelling higher-order characteristics adds variance to
the synthetic data that induces an increase in the \(pMSE\).
Additionally, Table~\ref{tbl-sim2-results} shows that the performance of
all utility measures improves with the sample size, and decreases when
the dimensionality of the data increases. The only exception to the
latter trend is the logistic regression based \(pMSE\)-ratio model,
which consistently improves with the dimensionality of the data.

\begin{figure}[t]

\centering{

\includegraphics[width=1\linewidth,height=\textheight,keepaspectratio]{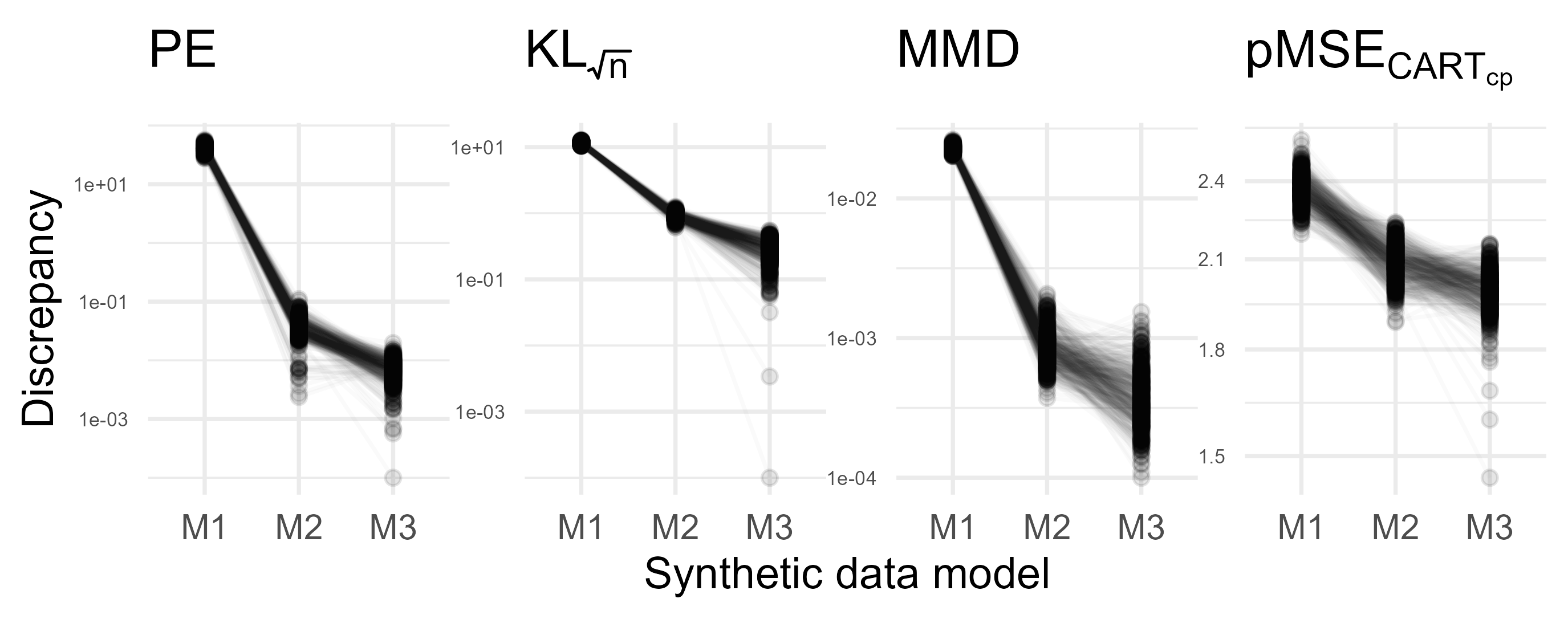}

}

\caption{\label{fig-sim2-results}The estimated utility measures (\(PE\),
\(KL_{\sqrt{n}}\), \(MMD\) and \(pMSE_{\text{cart}_{cp}}\)) for
\(n = 1000\) observations and \(D = 50\) variables in \(500\)
simulations for the three increasingly accurate synthetic data models.
All metrics show how the estimated utility improves with the refinements
of the synthesis models, but \(PE\) and \(KL_{\sqrt{n}}\) show less
variability and catch the synthesis model improvements more
consistently.}

\end{figure}

To zoom in on the behavior of four methods, we visualize the results of
default uLSIF, the KL-divergence based on \(k = \sqrt{n}\) neighbors,
the \(pMSE\)-ratio based on a CART model and the \(MMD\) for
\(n = 1000\) and \(D = 50\) in Figure~\ref{fig-sim2-results}. The
\(KL_{\sqrt{n}}\) consistently recognizes improvements in the synthesis
model. This also holds to a large extent for the Pearson divergence
(\(PE\)), where all but a few lines are parallel and in the correct
direction. The \(MMD\) becomes slightly more noisy, while for the
\(pMSE_{\text{cart}_{cp}}\), the variability in paths substantially
increases. Finally, the figures show that the improvement in utility due
to modelling the covariances rather than only the means and variances is
much larger than the improvement due to also modelling the non-linear
relationships.

\linespread{1}

\linespread{2}

\subsubsection{Results 2: Mixed-data
simulations}\label{results-2-mixed-data-simulations}

In the mixed data simulations, a similar pattern arises. For the
smallest sample size, the KL-divergence based on the nearest neighbor
performs remarkably well in this difficult setting, while all other
methods clearly struggle to find the correct ordering. For larger sample
sizes, the uLSIF-based Pearson divergence (\(PE\) and
\(PE_\text{fixed}\)) and \(k\)-nearest neighbors-based KL-divergence and
the \(pMSE\)-ratio with main effects and interactions
(\(pMSE_\text{logit, int}\)) perform essentially on par and detect the
correct ordering in close to every simulation. Again, the density ratio
models are relatively robust to model specification, although KLIEP
performs substantially worse than uLSIF except for the smallest sample
size (Appendix \ref{sec-app-multivariate},
Table~\ref{tbl-app-multivariate}). Note that in this simulation, the
\(pMSE\)-ratio with main effects and interactions contains the terms
that distinguish the three synthesis models, and thus performs much
better than in the previous simulations, but cannot be estimated with
only \(100\) samples.

\linespread{1}

\begin{table}

\caption{\label{tbl-sim3-results}Proportion of simulations (Setting 2)
in which the true synthetic data model order is captured by the utility
measure, for default uLSIF (\(PE\)), with fixed hyperparameters
(\(PE_\text{fixed}\)), KL-divergence based on its nearest neighbor
(\(KL_1\)) and based on its \(\sqrt{n}\)th nearest neighbor, maximum
mean discrepancy (\(MMD\)) and \(pMSE\)-ratio based on a CART model with
small complexity parameter (\(pMSE_{\text{CART}_{cp}}\)), based on a
logistic model with main effects (\(pMSE_\text{logit}\)) and a logit
model with main and interaction effects
(\(pMSE_\text{logit, int}\)).\vspace{0.4cm}}

\centering{

\begin{tabularx}{\linewidth}{>{\raggedright\arraybackslash}X>{\raggedright\arraybackslash}p{10mm}>{\centering\arraybackslash}p{10mm}>{\centering\arraybackslash}p{10mm}>{\centering\arraybackslash}p{10mm}>{\centering\arraybackslash}p{10mm}>{\centering\arraybackslash}p{17mm}>{\centering\arraybackslash}p{17mm}>{\centering\arraybackslash}p{17mm}}
\toprule
$n$ & $PE$ & $PE_{\text{fixed}}$ & $KL_1$ & $KL_{\sqrt{n}}$ & $MMD$ & $pMSE_{\text{CART}_{cp}}$ & $pMSE_{\text{logit}}$ & $pMSE_{\text{logit, int}}$\\
\midrule
100 & 0.14 & 0.57 & 0.73 & 0.51 & 0.30 & 0.24 & 0.14 & -\\
1000 & 0.99 & 0.99 & 0.97 & 0.99 & 0.72 & 0.59 & 0.13 & 0.99\\
2000 & 1.00 & 1.00 & 1.00 & 1.00 & 0.86 & 0.71 & 0.14 & 1.00\\
\bottomrule
\end{tabularx}

}

\end{table}%

\linespread{2}

These results have a remarkable implication for the evaluation of
synthetic data generation methods. When the aim is to rank candidate
synthetic data sets using a global utility statistic, using a rather
flexible model seems advisable. In Section~\ref{sec-univ-sims} and
Appendix \ref{sec-app-B}, we showed that the \(k\)-nearest neighbor
density ratio estimate is substantially more variable than the
uLSIF-based estimate, but when the aim is to quantify the utility of the
synthetic data, this variability does not seem to be problematic. Also,
uLSIF with fixed hyperparameters (with a small bandwidth and small
regularization parameter, thus rendering a more flexible model) yields a
correct ordering more often than uLSIF with optimized hyperparameters,
and a CART model with smaller complexity parameter performs better than
a CART model with the default settings (Appendix
\ref{sec-app-multivariate}). These simulations suggest that a more
flexible model is better able to detect discrepancies between the
observed and synthetic data. Obtaining a stable and accurate estimate of
the density ratio (or, by analogy, of the propensity scores in the
\(pMSE\)) does not seem to be required when one is solely interested in
globally ranking different synthetic data sets. However, accurately
obtaining local information on the misfit of the synthetic data is
valuable in itself, which we illustrate in the next section.

\section{Application: Synthetic data generation for the U.S. Current
Population Survey}\label{sec-app}

\linespread{1}

\linespread{2}

We apply density ratio estimation to evaluate the utility of synthetic
versions of the March 2000 U.S. Current Population Survey (CPS),
building on the setup by Drechsler
(\citeproc{ref-drechsler_utility_2022}{2022}). That is, we focus on the
continuous variables age, household income, household property taxes and
social security payments, and the categorical variables sex, race,
marital status and education level measured on a subsample of
\(n_\text{obs} = 5000\) individuals. Note that all continuous variables
except age are substantially skewed. Moreover, the variables household
property taxes and social security payments have a point-mass at zero
(see the first column in Figure~\ref{fig-application-distributions}). To
account for the skewness, we transform the continuous variables to
cube-root scale (\(f(x) = |x|^{1/3}\cdot\text{sign}(x)\)). We compare
two different synthesis strategies that differ in how they treat the
point-mass at zero (using the \texttt{R}-package \texttt{synthpop};
\citeproc{ref-synthpop}{Nowok, Raab, and Dibben 2016}). The first
modelling strategy applies a linear regression model on the transformed
continuous variables. The categorical variables are modelled using
(multinomial) logistic regression models. The second modelling strategy
extends the first strategy by applying a semi-continuous model to the
transformed variables household property taxes and social security
payments. Here, the point mass is modelled first using a logistic
regression model, after which the non-zero values are synthesized using
a linear regression model. All other variables are modelled according to
the first strategy. We generate \(m = 5\) synthetic data sets for both
strategies, each with \(n_\text{syn} = 5000\) observations, and evaluate
their utility using uLSIF with the same default settings as in the
previous sections.

\subsection{Evaluating global utility using density ratio
estimation}\label{evaluating-global-utility-using-density-ratio-estimation}

\linespread{1}

\begin{figure}[t]

\centering{

\includegraphics[width=1\linewidth,height=\textheight,keepaspectratio]{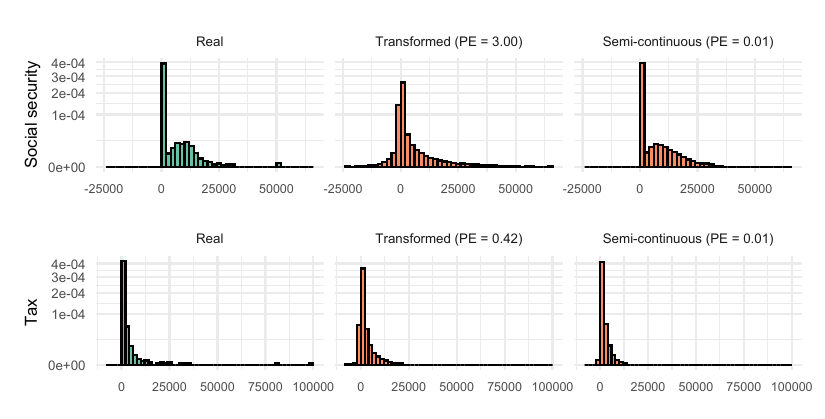}

}

\caption{\label{fig-application-distributions}Real and synthetic data
distributions for the variables social security payments and household
property taxes for the transformed and semi-continuous synthesis
strategies (averaged over five synthetic data sets). The panel titles
display the estimated Pearson divergence for each variable between the
observed and synthetic data, estimated using a density ratio model for
each variable separately. Note that the y-axis is displayed on a
square-root scale to improve visibility.}

\end{figure}%

\linespread{2}

Figure~\ref{fig-application-distributions} displays the distributions of
the variables social security payments (Social security) and household
property taxes (Tax) for the real data and the synthetic data according
to the transformed and semi-continuous modelling strategies. The figure
shows that the semi-continuous modelling strategy fits the observed data
better than the transformed modelling strategy, as it better captures
the point mass and does not yield an excessive number of negative
values. Note that there are only minor differences for the other
variables, as there are no differences in synthesis methods. When
evaluating the global utility of the two synthesis strategies, the
improvement in synthesis models is reflected in the Pearson divergence.
That is, the transformed modelling strategy has a higher average
estimated Pearson divergence
(\(\overline{PE}_\text{transformed} = 0.405\),
\(\text{range} = [0.318, 0.452]\)) than the semi-continuous modelling
strategy (\(\overline{PE}_\text{semi-continuous} = 0.104\),
\(\text{range} = [0.093, 0.128]\)). Although the difference may seem
small in an absolute sense, it is substantial in a relative sense.
Additionally, the variability in estimated Pearson divergences for each
synthesis strategy is much smaller than the difference between the two
strategies. Finally, we remark that when estimating the density ratio on
the level of the individual variables, the semi-continuous modelling
strategy obtains substantially lower Pearson divergences than the
transformed modelling strategy (see the panel titles in
Figure~\ref{fig-application-distributions}). Hence, we conclude that the
Pearson divergence serves as an adequate measure of synthetic data
utility.

\subsection{Density ratios as local utility
estimates}\label{density-ratios-as-local-utility-estimates}

\linespread{1}

\begin{figure}[t]

\centering{

\includegraphics[width=1\linewidth,height=\textheight,keepaspectratio]{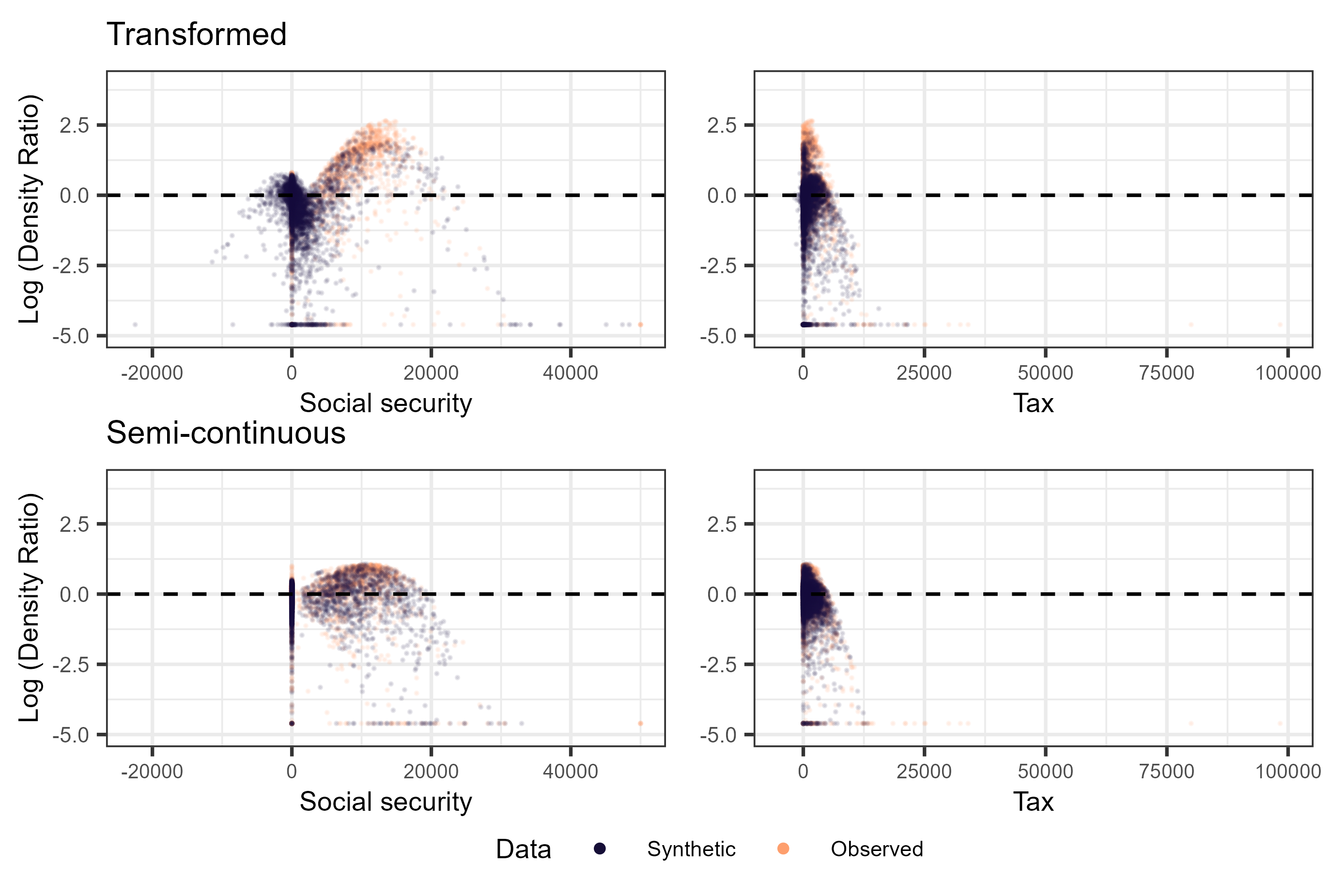}

}

\caption{\label{fig-app-utility-results}Scatter plots for social
security payments (Social security) and household property taxes (Tax)
versus the pointwise estimated density ratio for each record (on a
log-scale) for the first synthetic data set under both synthesis
strategies and the observed data. Note that the density ratio is
estimated using all variables; vertical spread reflects differences due
to other variables.}

\end{figure}%

\linespread{2}

Besides quantifying the global utility of synthetic data, the density
ratio also provides local information about the utility of synthetic
data. The estimated density ratio can provide an intuition about which
variables are not modelled adequately. That is, by plotting the
estimated density ratio against each variable or pairs of variables, one
can identify variables for which there is a clear pattern in the density
ratio values. Figure~\ref{fig-app-utility-results} shows the estimated
density ratio values (estimated over all variables, on a log-scale)
plotted against the variables social security payments and household
property taxes for the first synthetic data set under both synthesis
strategies. First, the figure shows that the spread in density ratio
values is substantially larger for the transformed modelling strategy
than for the semi-continuous modelling strategy. Second, the density
ratio values indicate that the transformed modelling strategy misses the
point masses at zero, as indicated by the small bumps located at zero
for both point masses. For social security values slightly larger than
zero, too many synthetic records were generated, as indicated by the
black band of points at the bottom of the top left panel (estimates
below 0.01 are truncated). Finally, the larger bump around \(12000\)
indicates that far too few synthetic records are generated in this
region. For household property taxes, the estimated density ratio values
overlap more for both synthesis strategies, showing that at least
marginally, the discrepancy between the observed data distribution and
the synthetic data distributions is smaller.

For the semi-continuous modelling strategy, there is some spread in
density ratio values, but there is no clear pattern, and the synthetic
and observed data points are mostly overlapping. Hence, marginally, the
density ratio values do not reveal a substantial misfit of the
semi-continuous strategy. Additionally, we note that in the tails of the
distribution, there is little information to estimate the density ratio
accurately, and the estimates are shrunk towards zero. From a utility
perspective, this might be undesirable, as misfit in the tails of the
distribution might go by unnoticed. However, from a privacy perspective,
this regularization is beneficial, as it prevents the density ratio from
becoming very large around the location of outlying observed data
points, potentially revealing the location of these points. The
regularization thus limits the influence of individual observations.

\section{Discussion and conclusion}\label{discussion-and-conclusion}

High quality utility measures are essential when creating synthetic data
for use in downstream tasks. In this paper, we provided a framework for
evaluating synthetic data utility through density ratio estimation.
Viewing synthetic data utility through the lens of density ratio
estimation yields several appealing properties. The density ratio
perspective encompasses global and local utility in a single framework.
Divergence measures based on the density ratio provide insight into the
quality of the synthetic data, while the density ratio itself can be
used to guide the synthesis process. Our simulations and application
show that the density ratio estimation framework can be used in a wide
variety of settings. In the multivariate simulations, density
ratio-based methods, such as unconstrained least-squares importance
fitting (uLSIF) and the nearest neighbor-based KL-divergence, commonly
ranked synthetic data sets more reliably than the \(pMSE\)-ratio and the
maximum mean discrepancy. In the application, we showed how the density
ratio itself provides information on how the synthesis model can be
improved. Beyond the density ratio estimation methods discussed, several
extensions are available to deal with high-dimensional data, such as
spectral density ratio estimation
(\citeproc{ref-izbicki_dre_2014}{Izbicki, Lee, and Schafer 2014}) and
least-squares heterodistributional subspace search
(\citeproc{ref-sugiyama_lhss_2011}{Sugiyama, Yamada, et al. 2011}).
These extensions, and the density ratio estimation methods discussed in
this paper, are made accessible through the \texttt{densityratio}
package in \texttt{R} (\citeproc{ref-densityratio}{Volker 2023}).

Throughout the paper, we have contrasted the density ratio framework
with existing utility measures such as the \(pMSE\) and the
Kullback-Leibler divergence. Both measures can be considered special
cases of the density ratio framework. The Kullback-Leibler divergence
explicitly contains the density ratio in its definition, and Appendix
\ref{sec-app-B} describes how the \(k\)-nearest neighbor approach to
KL-divergence estimation gives rise to an estimate of the density ratio.
The \(pMSE\) can also be seen as a special case of the density ratio
framework, by noting that the predicted probabilities are easily
transformed into density ratios (see
\citeproc{ref-sugiyama_suzuki_kanamori_2012}{Sugiyama, Suzuki, and
Kanamori 2012a}). In fact, Menon and Ong
(\citeproc{ref-menon2016dreloss}{2016}) show that density ratio
estimation is equivalent to class probability estimation (as performed
in the \(pMSE\)), in the sense that minimizing a class probability
estimation loss is equivalent to minimizing a Bregman divergence to the
true density ratio function. These connections raise the question of
which model class, and which loss function, is most appropriate when the
goal is to evaluate the utility of synthetic data.

While we leave a formal evaluation of this question to future work, we
can provide some intuitions. First, our simulations showed that more
flexible methods, such as uLSIF with fixed hyperparameters, nearest
neighbor-based density ratio estimation and CART-based \(pMSE\) models
with smaller complexity parameters rank synthetic data sets more
accurately in terms of utility than their more stable variants in the
multivariate simulations. These findings suggest that rather variable
estimates of the density ratio (or the class probabilities) might not
harm the estimated divergence measures, because averaging over these
variable estimates provides stability by itself. At the same time,
utility measures ought to be robust to extreme values in the observed
data, in the sense that they should not depend too strongly on whether
similar extreme observations are reproduced in the synthetic data. After
all, extreme cases typically bear higher disclosure risk as they do not
blend into the crowd (\citeproc{ref-gehrke_crowd_2012}{Gehrke et al.
2012}). If the utility measure is too sensitive to these extreme cases,
small discrepancies that serve to protect privacy may seem to reflect
low quality synthetic data. Sugiyama, Suzuki, and Kanamori
(\citeproc{ref-sugiyama_bregman_2012}{2012b}) showed that uLSIF is
relatively robust to outliers compared to density ratio estimation using
the KL-loss. Relative uLSIF is more robust than the regular variant of
uLSIF (\citeproc{ref-yamada_rulsif_2013}{Yamada et al. 2013}), because
the relative density ratio is bounded from above, stabilizing the
resulting divergence when the observed data distribution has mass where
the synthetic data distribution has none. In our simulations, relative
uLSIF performed similarly to uLSIF. The robustness of density ratio
estimation techniques can be further improved by using more robust loss
functions, such as Basu's power divergence
(\citeproc{ref-basu_power_1998}{Basu et al. 1998};
\citeproc{ref-sugiyama_bregman_2012}{Sugiyama, Suzuki, and Kanamori
2012b}).

Additionally, utility measures should work well in a wide variety of
settings, so users do not need to fine-tune utility measures and can
focus on the problem at hand. To this end, non-parametric methods may be
preferred over parametric methods, as they do not hinge upon an
appropriate model specification. Although correctly specified parametric
models are more efficient in a statistical sense, they are less reliable
when the model is misspecified
(\citeproc{ref-sugiyama_suzuki_kanamori_2012}{Sugiyama, Suzuki, and
Kanamori 2012a}). Our results suggest that uLSIF serves as a suitable
compromise between flexibility and robustness, although the
KL-divergence using \(k\)-nearest neighbor density ratio estimation may
be more sensitive to discrepancies between the observed and synthetic
data. The KL importance estimation procedure (KLIEP) was less suited as
a utility measure: uLSIF proved more reliable in capturing the ranking
of synthetic data sets, and is computationally much more efficient.

Despite the appealing characteristics of density ratio estimation,
several open questions remain. Kernel methods, as typically employed in
density ratio estimation, can be sensitive to the choice of kernel, and
the use of a Gaussian kernel, while a reasonable and versatile default,
may not be appropriate in every situation. Categorical variables, for
instance, enter the Gaussian kernel through one indicator per category.
Our simulations and application suggest that with sufficient data, this
does not pose severe problems to the density ratio estimates. However,
when the number of categories is close to the number of observations,
obtaining an accurate estimate of the density ratio is hard, and all
methods except the nearest neighbors-based KL estimator with \(k=1\)
performed poorly. Using distance measures or kernels that are tailored
to categorical data might yield improvements when using standard density
ratio estimation techniques. Similar problems may occur when the number
of variables is large relative to the number of observations, in which
case density ratio estimation methods tailored to high-dimensional data
may bring relief (\citeproc{ref-izbicki_dre_2014}{Izbicki, Lee, and
Schafer 2014}; \citeproc{ref-sugiyama_lhss_2011}{Sugiyama, Yamada, et
al. 2011}). Since essentially all methods performed better when the
amount of data increases, it might be interesting to explore whether
simply generating a larger amount of synthetic data would already be
sufficient to reap some of the benefits of having more data. While
obtaining a larger number of real observations is typically costly or
infeasible for individual researchers, obtaining a larger number of
synthetic samples is relatively cheap. Using these synthetic samples
reduces one source of variability when comparing different synthetic
data sets, and may thus provide more informative utility estimates.

Finally, we remark that density ratio estimates can be used directly to
calibrate inferences with synthetic data to the true data
(\citeproc{ref-ghalebikesabi_dpiw_2022}{Ghalebikesabi et al. 2022}),
potentially reducing bias in statistical analyses with synthetic data.
However, such an approach requires differentially private estimates of
the density ratio. Since the weights are estimated from the observed
data, they might leak sensitive information. For instance, large weights
in sparse regions may reveal the location of real records if the weights
were released publicly. We note that while the estimated density ratios
might be sensitive in themselves, there is typically no need to publish
these. Utility evaluation is typically an internal procedure, performed
by the data provider, that is used to determine which data set to
publish. Future work could evaluate whether density ratio estimation can
also be carried out in a differentially private way, such that
importance weights can be safely constructed and disseminated. The
density ratio has also been proposed as a privacy metric in its own
right, allowing the data provider to detect overfitting of the synthesis
model by locating small clusters where substantially more synthetic
samples occur than one would expect based on an independent reference
sample from the observed data distribution
(\citeproc{ref-vanbreugel_membership_2023}{Breugel et al. 2023}). It
might be worthwhile to further explore the connections between density
ratio estimation for detecting privacy risks and for utility evaluation.
In principle, synthetic data utility should never be judged on its own,
because utility measures reward synthetic data for being as similar to
the observed data as possible. Balancing privacy with utility in a
common framework might therefore prove fruitful.

Despite these open questions, we have shown that already in its current
form, density ratio estimation provides a flexible and informative
framework for evaluating synthetic data utility. Highly flexible
specifications typically rank synthetic data sets well in terms of
utility, especially when sample sizes are small. Automatically tuned
uLSIF provides a more stable estimate of the density ratio and requires
essentially no specification on behalf of the user, while still working
well in a wide variety of settings. The framework can be readily
incorporated into synthetic data workflows using the
\texttt{densityratio} package.

\section*{Acknowledgements}\label{acknowledgements}
\addcontentsline{toc}{section}{Acknowledgements}

We are grateful to Joerg Drechsler for sharing the U.S. Current
Population Survey data, and to Stef van Buuren, Gerko Vink, Hanne
Oberman and Carlos Gonzalez Poses for many valuable discussions. We also
thank three anonymous reviewers for comments that led to substantial
improvements of this manuscript.

\section*{References}\label{references}
\addcontentsline{toc}{section}{References}

\phantomsection\label{refs}

\setcounter{section}{0}
\renewcommand{\thesection}{\Alph{section}}

\setcounter{table}{0}
\renewcommand{\thetable}{A\arabic{table}}

\setcounter{figure}{0}
\renewcommand{\thefigure}{A\arabic{figure}}

\section{\texorpdfstring{Density ratio estimation in
\texttt{R}}{Density ratio estimation in R}}\label{sec-app-A}

To perform density ratio estimation in \texttt{R}, we can use the
\texttt{densityratio} package, which includes model fitting functions
for multiple popular loss functions. Additionally, we can complement the
package with standard optimization routines in \texttt{R}. Here, we show
an example of both approaches.

We generate two small toy data sets, \texttt{x} and \texttt{y}, that
serve as numerator and denominator data, respectively. Moreover, we set
\texttt{x} as the centers in the Gaussian kernel, and we pre-define the
kernel bandwidth to be \(\sigma = 2\) and fix the regularization
parameter to \(\lambda = 0.5\). Note that we do not perform
cross-validation for hyperparameter selection to enhance illustrational
clarity. Subsequently, we fit a density ratio model using the
\texttt{ulsif()} function from the \texttt{densityratio} package.

\linespread{1}

\begin{Shaded}
\begin{Highlighting}[]
\FunctionTok{library}\NormalTok{(densityratio)}

\FunctionTok{set.seed}\NormalTok{(}\DecValTok{123}\NormalTok{)}

\NormalTok{x }\OtherTok{\textless{}{-}} \FunctionTok{rnorm}\NormalTok{(}\DecValTok{250}\NormalTok{) }\SpecialCharTok{|\textgreater{}} \FunctionTok{as.matrix}\NormalTok{()}
\NormalTok{y }\OtherTok{\textless{}{-}} \FunctionTok{rnorm}\NormalTok{(}\DecValTok{250}\NormalTok{, }\DecValTok{1}\NormalTok{, }\FunctionTok{sqrt}\NormalTok{(}\DecValTok{2}\NormalTok{)) }\SpecialCharTok{|\textgreater{}} \FunctionTok{as.matrix}\NormalTok{()}

\NormalTok{fit\_dr }\OtherTok{\textless{}{-}} \FunctionTok{ulsif}\NormalTok{(x, y, }\AttributeTok{centers =}\NormalTok{ x, }\AttributeTok{scale =} \ConstantTok{NULL}\NormalTok{, }\AttributeTok{sigma =} \DecValTok{2}\NormalTok{, }\AttributeTok{lambda =} \FloatTok{0.5}\NormalTok{)}
\NormalTok{rhat\_dr }\OtherTok{\textless{}{-}} \FunctionTok{predict}\NormalTok{(fit\_dr)}
\end{Highlighting}
\end{Shaded}

\linespread{2}

We can obtain the same solution using standard optimization techniques.
In order to do this, we first need to define the kernel matrices for the
numerator and denominator data, and define a loss function that can be
used in optimization. We then use the \texttt{optim()} function to
minimize the loss function. Recall that we use a linear model for the
density ratio, as per Equation~\ref{eq-dr-model}, we use a Gaussian
kernel as basis function, and we use the least-squares loss of
Equation~\ref{eq-squared-error-loss}.

\linespread{1}

\begin{Shaded}
\begin{Highlighting}[]
\NormalTok{Dx }\OtherTok{\textless{}{-}}\NormalTok{ densityratio}\SpecialCharTok{:::}\FunctionTok{distance}\NormalTok{(x, x, }\ConstantTok{FALSE}\NormalTok{)}
\NormalTok{Dy }\OtherTok{\textless{}{-}}\NormalTok{ densityratio}\SpecialCharTok{:::}\FunctionTok{distance}\NormalTok{(y, x, }\ConstantTok{FALSE}\NormalTok{)}

\NormalTok{Kx }\OtherTok{\textless{}{-}}\NormalTok{ densityratio}\SpecialCharTok{:::}\FunctionTok{kernel\_gaussian}\NormalTok{(Dx, }\AttributeTok{sigma =} \DecValTok{2}\NormalTok{)}
\NormalTok{Ky }\OtherTok{\textless{}{-}}\NormalTok{ densityratio}\SpecialCharTok{:::}\FunctionTok{kernel\_gaussian}\NormalTok{(Dy, }\AttributeTok{sigma =} \DecValTok{2}\NormalTok{)}

\NormalTok{loss }\OtherTok{\textless{}{-}} \ControlFlowTok{function}\NormalTok{(alpha, Kx, Ky) \{}
  \FunctionTok{mean}\NormalTok{((Ky }\SpecialCharTok{\%*\%}\NormalTok{ alpha }\SpecialCharTok{{-}} \DecValTok{1}\NormalTok{) }\SpecialCharTok{*}\NormalTok{ Ky }\SpecialCharTok{\%*\%}\NormalTok{ alpha }\SpecialCharTok{{-}}\NormalTok{ (Ky }\SpecialCharTok{\%*\%}\NormalTok{ alpha }\SpecialCharTok{{-}} \DecValTok{1}\NormalTok{)}\SpecialCharTok{\^{}}\DecValTok{2} \SpecialCharTok{/} \DecValTok{2}\NormalTok{) }\SpecialCharTok{{-}}
    \FunctionTok{mean}\NormalTok{(Kx }\SpecialCharTok{\%*\%}\NormalTok{ alpha }\SpecialCharTok{{-}} \DecValTok{1}\NormalTok{) }\SpecialCharTok{+}
    \FloatTok{0.5}\SpecialCharTok{/}\DecValTok{2} \SpecialCharTok{*} \FunctionTok{sum}\NormalTok{(alpha}\SpecialCharTok{\^{}}\DecValTok{2}\NormalTok{) }\CommentTok{\# ridge penalty}
\NormalTok{\}}

\NormalTok{alpha }\OtherTok{\textless{}{-}} \FunctionTok{runif}\NormalTok{(}\FunctionTok{ncol}\NormalTok{(Kx))}
\NormalTok{fit\_optim }\OtherTok{\textless{}{-}} \FunctionTok{optim}\NormalTok{(alpha, loss, }\AttributeTok{Kx =}\NormalTok{ Kx, }\AttributeTok{Ky =}\NormalTok{ Ky, }
                   \AttributeTok{method =} \StringTok{"BFGS"}\NormalTok{, }
                   \AttributeTok{control =} \FunctionTok{list}\NormalTok{(}\AttributeTok{maxit =} \DecValTok{100}\NormalTok{))}

\NormalTok{rhat\_optim }\OtherTok{\textless{}{-}}\NormalTok{ Kx }\SpecialCharTok{\%*\%}\NormalTok{ fit\_optim}\SpecialCharTok{$}\NormalTok{par}
\end{Highlighting}
\end{Shaded}

\linespread{2}

\begin{figure}[t]

\centering{

\pandocbounded{\includegraphics[keepaspectratio]{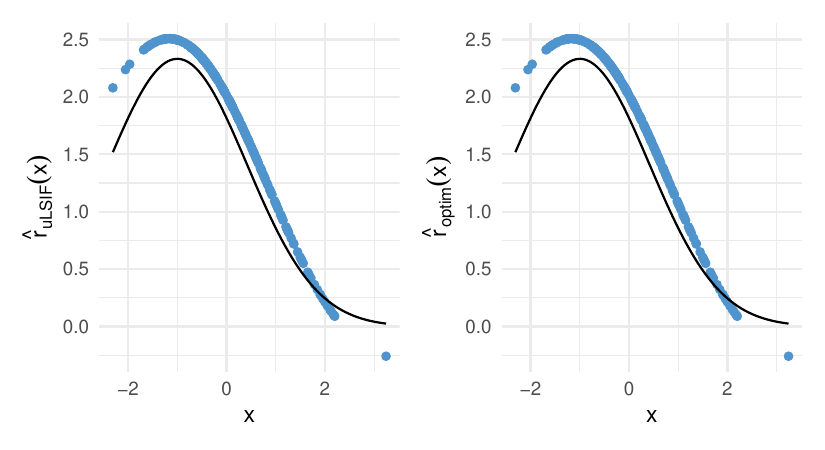}}

}

\caption{\label{fig-AppA}Estimated density ratio values at the locations
of the numerator samples (the blue points) compared to the true density
ratio function, for both the \texttt{ulsif()} function from the
\texttt{densityratio} package and standard optimization in \texttt{R}
using the BFGS algorithm.}

\end{figure}

Subsequently, we can show that both procedures yield virtually identical
results (see Figure~\ref{fig-AppA}). Moreover, the sum of squared
difference between the estimated density ratio parameters equals
1.54e-12, and the sum of squared difference between the predicted
density ratios equals 3.8e-10.

\linespread{1}

\linespread{2}

If we were to perform cross-validation for hyperparameter selection, we
could run the \texttt{ulsif()}-function without specifying the
\texttt{sigma} and \texttt{lambda} arguments. The function then performs
a grid search over a predefined range of hyperparameters. Moreover, it
is possible to supply a vector of \texttt{sigma} and \texttt{lambda}
values to \texttt{ulsif()}, after which cross-validation is
automatically performed.

\linespread{1}

\begin{Shaded}
\begin{Highlighting}[]
\NormalTok{fit }\OtherTok{\textless{}{-}} \FunctionTok{ulsif}\NormalTok{(x, y, }\AttributeTok{centers =}\NormalTok{ x, }\AttributeTok{scale =} \ConstantTok{NULL}\NormalTok{)}
\NormalTok{fit}
\end{Highlighting}
\end{Shaded}

\begin{verbatim}

Call:
ulsif(df_numerator = x, df_denominator = y, scale = NULL, centers = x)

Kernel Information:
  Kernel type: Gaussian with L2 norm distances
  Number of kernels: 250
  sigma: num [1:10] 0.0601 0.177 0.2971 0.4228 0.5583 ...

Regularization parameter (lambda): num [1:20] 1000 483.3 233.6 112.9 54.6 ...

Optimal sigma (loocv): 0.7045195
Optimal lambda (loocv): 1.43845
Optimal kernel weights (loocv): num [1:250] 0.03193 0.02459 0.00951 0.01312 0.01078 ...
 
\end{verbatim}

\linespread{2}

The model output shows the optimal \texttt{sigma} parameter out of a
vector containing ten values, and the optimal \texttt{lambda} parameter
out of a vector containing twenty values. The package also includes
density ratio estimation techniques, such as \texttt{kliep()},
\texttt{kmm()}, \texttt{spectral()}, \texttt{lhss()} and
\texttt{naive()}, that all work in a similar fashion. Additionally, the
package contains generic functions to \texttt{predict()} with,
\texttt{plot()}, or summarize (e.g., \texttt{summary()}) the fitted
models. Moreover, computationally intensive code is implemented in
\texttt{C++} through \texttt{Rcpp} (\citeproc{ref-Rcpp}{Eddelbuettel
2013}) and most functions allow for parallel computation through the
\texttt{openmp}-backend in \texttt{C++}. The interested reader is
referred to the package documentation
(\citeproc{ref-densityratio}{Volker 2023}) for additional information.

\section{\texorpdfstring{\(k\)-Nearest neighbor density ratio
estimation}{k-Nearest neighbor density ratio estimation}}\label{sec-app-B}

For a record \(\bx_i\) in the observed data, the \(k\)-nearest neighbor
estimates of the observed and synthetic data densities are \[
\hat{p}_\text{obs}(\bx) = \frac{k}{\nobs-1} \cdot \frac{\Gamma(D/2 + 1)}{\pi^{D/2} \rho_k(\bx_i)^D}
\] and \[
\hat{p}_\text{syn}(\bx) = \frac{k}{\nsyn} \cdot \frac{\Gamma(D/2 + 1)}{\pi^{D/2} \nu_k(\bx_i)^D},
\] where \(\rho_k(\bx_i)\) and \(\nu_k(\bx_i)\) are the Euclidean
distance from \(\bx_i\) to its \(k\)-th nearest neighbor among the
observed and synthetic records, respectively, and
\(\pi^{D/2} / \Gamma(D/2 + 1)\) is the volume of the \(D\)-dimensional
unit ball (see, e.g., \citeproc{ref-wang_divergence_2009}{Wang,
Kulkarni, and Verdu 2009}). Taking their ratio yields \[
\hat{r}(\bx_i) = \frac{\nsyn}{\nobs - 1} \left( \frac{\nu_k(\bx_i)}{\rho_k(\bx_i)} \right)^D,
\] where the volume terms are identical and thus cancel. The
\(k\)-nearest neighbor estimator of the Kullback-Leibler divergence is
the average of the log density ratio over the observed records, \[
\begin{aligned}
\hat{KL}_{knn} &= \frac{1}{\nobs} \sum_{i=1}^{\nobs} \log \hat{r}(\bx_i) \\
&= \frac{D}{\nobs} \sum_{i=1}^{\nobs} \log \frac{\nu_k(\bx_i)}{\rho_k(\bx_i)} + \log \frac{\nsyn}{\nobs - 1},
\end{aligned}
\] which implicitly estimates the ratio.

Applying this estimate of the density ratio to the simulated data from
the univariate simulations, and setting \(k = 15 \approx \sqrt n\), as
suggested by the authors who introduced \(k\)-nearest neighbor density
estimation (\citeproc{ref-loftsgaarden_nonparametric_1965}{Loftsgaarden
and Quesenberry 1965}), we can compare the performance of the
\(k\)-nearest neighbor density ratio estimation to the density ratio
estimation through uLSIF.

\linespread{1}

\begin{figure}[t]

\centering{

\includegraphics[width=1\linewidth,height=\textheight,keepaspectratio]{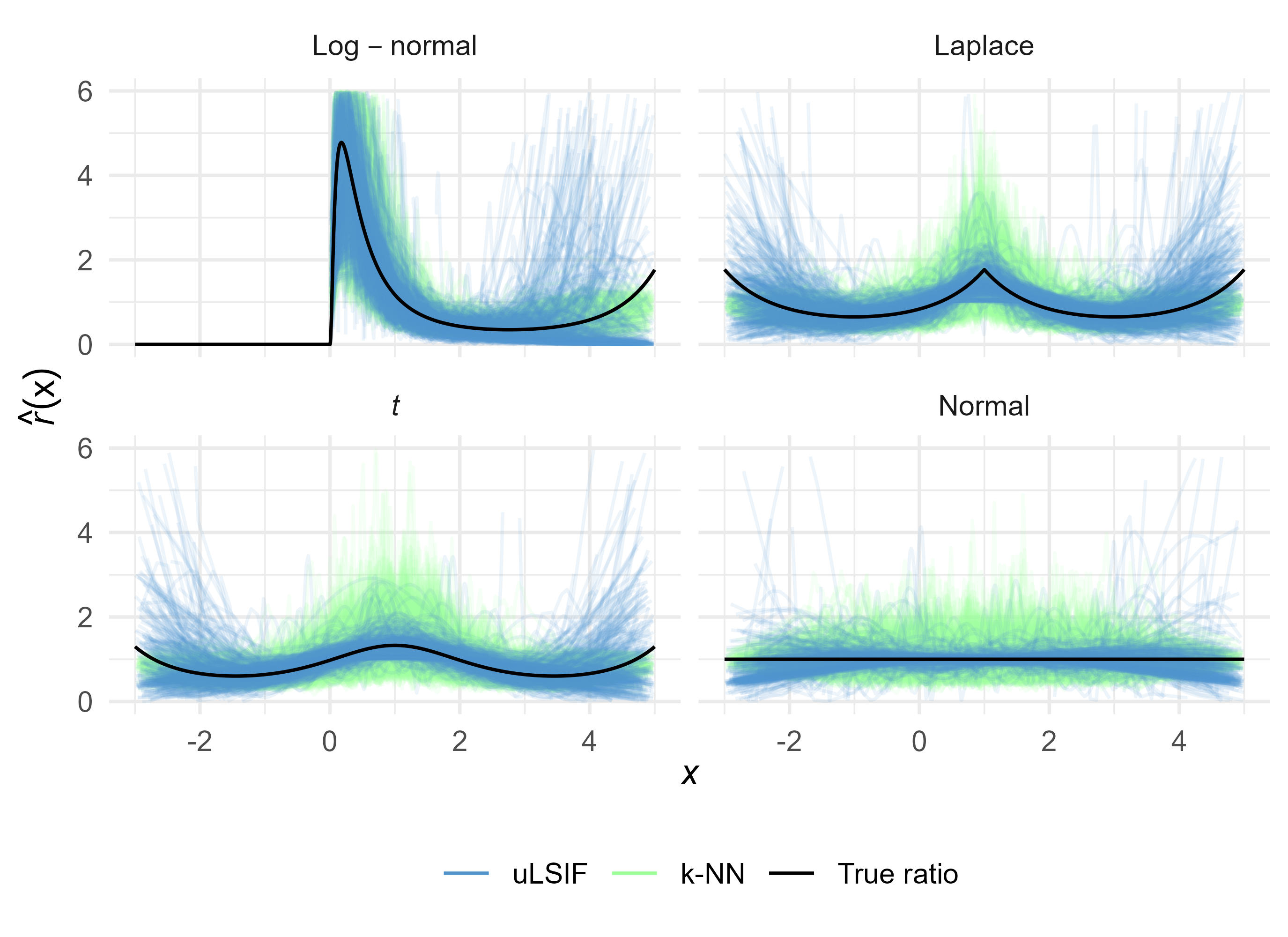}

}

\caption{\label{fig-AppB}Density ratio estimation through \(k\)-nearest
neighbor density ratio estimation (\(k = 15\)) and uLSIF based on 500
replications. The black line depicts the true density ratio function..}

\end{figure}%

\linespread{2}

The \(k\)-nearest neighbor density ratio estimator performs well on
average (the green lines in Figure~\ref{fig-AppB}), especially in the
center of the distribution. However, it suffers from bias in the tails,
where the number of observations is low and the density ratio is
underestimated. Moreover, \(k\)-nearest neighbor density ratio
estimation has much higher variance in the center of the distribution
than density ratio estimation through uLSIF, as can be seen by the fact
that the green lines cover a much wider band than the blue lines. In the
tails of the distribution, uLSIF is substantially more variable, but
since there are only few cases in these regions, the effect on the
resulting divergence estimate is relatively small.

\section{Stability of density ratio estimation under different model
specifications}\label{sec-app-stability}

To evaluate the stability of density ratio estimation to the specifics
of the exact model, we generate multiple synthetic data sets that are
increasingly different from the true data from
Section~\ref{sec-univ-sims}. We use the four univariate data-generating
mechanisms from Section~\ref{sec-univ-sims} (log-normal, Laplace,
location-scale \(t\) and normal), all with mean \(\mu = 1\) and variance
\(\sigma^2 = 2\). In each simulation run, we draw an observed data set
of size \(\nobs \in \{250, 500\}\) from the true distribution, and five
synthetic data sets of size \(\nsyn = \nobs\) from the mixture
\((1 - p) f_{\text{true}} + p\,\mathcal{N}(1, 2)\), with
\(p \in \{0, 0.25, 0.5, 0.75, 1\}\). That is, each synthetic record is
drawn from the approximating normal distribution with probability \(p\),
and from the true distribution otherwise. Hence, the divergence between
the observed and synthetic data increases with the mixing probability
\(p\). For the normal mechanism, the mixture equals the true
distribution for every \(p\), so this scenario serves as a null case. We
repeat this 500 times for each mechanism and sample size.

For each synthetic data set, we compute

\begin{enumerate}
\def\labelenumi{\arabic{enumi}.}
\tightlist
\item
  \texttt{PE\_opt}: the Pearson divergence estimated with uLSIF using
  the default settings (200 kernel centers, with the bandwidth and
  regularization parameter selected by leave-one-out cross-validation);
\item
  \texttt{PE\_ce}: the same estimator with 100 kernel centers;
\item
  \texttt{PE\_fixed}: the same estimator with fixed hyperparameters,
  using the \(0.2\)-quantile distance between the observations and the
  kernel centers as bandwidth and \(\lambda = 1/\nobs\);
\item
  \texttt{PE\_rel02}: the relative Pearson divergence estimated with
  relative uLSIF (\citeproc{ref-yamada_rulsif_2013}{Yamada et al.
  2013}), with \(\alpha = 0.2\);
\item
  \texttt{PE\_rel05}: the relative Pearson divergence estimated with
  relative uLSIF (\citeproc{ref-yamada_rulsif_2013}{Yamada et al.
  2013}), with \(\alpha = 0.5\);
\item
  \texttt{KLIEP\_opt}: the Kullback-Leibler divergence estimated with
  KLIEP (\citeproc{ref-sugiyama_kliep_2007}{Sugiyama et al. 2007}), with
  the bandwidth selected by cross-validation;
\item
  \texttt{KLIEP\_fixed}: the Kullback-Leibler divergence estimated with
  KLIEP (\citeproc{ref-sugiyama_kliep_2007}{Sugiyama et al. 2007}), with
  the bandwidth fixed at the 0.2 quantile of the distances;
\item
  \texttt{KL\_knn}: the Kullback-Leibler divergence estimated with
  \(k\)-nearest neighbor density ratio estimation, with
  \(k = \lfloor \sqrt{\nobs} \rfloor\) (\citeproc{ref-kldest}{Hartung
  2024});
\item
  \texttt{MMD}: the unbiased estimate of the squared maximum mean
  discrepancy with a Gaussian kernel and the default bandwidth heuristic
  in kernlab (\citeproc{ref-gretton_kernel_2012}{Gretton et al. 2012};
  \citeproc{ref-kernlab}{Karatzoglou et al. 2004});
\item
  \texttt{sPMSE}: the \(pMSE\)-ratio based on a table with five groups
  (\citeproc{ref-raab2021assessing}{Raab, Nowok, and Dibben 2021}), the
  default for a single variable in \texttt{synthpop}
  (\citeproc{ref-synthpop}{Nowok, Raab, and Dibben 2016}).
\end{enumerate}

\linespread{1}

\begin{figure}[H]

\centering{

\includegraphics[width=1\linewidth,height=\textheight,keepaspectratio]{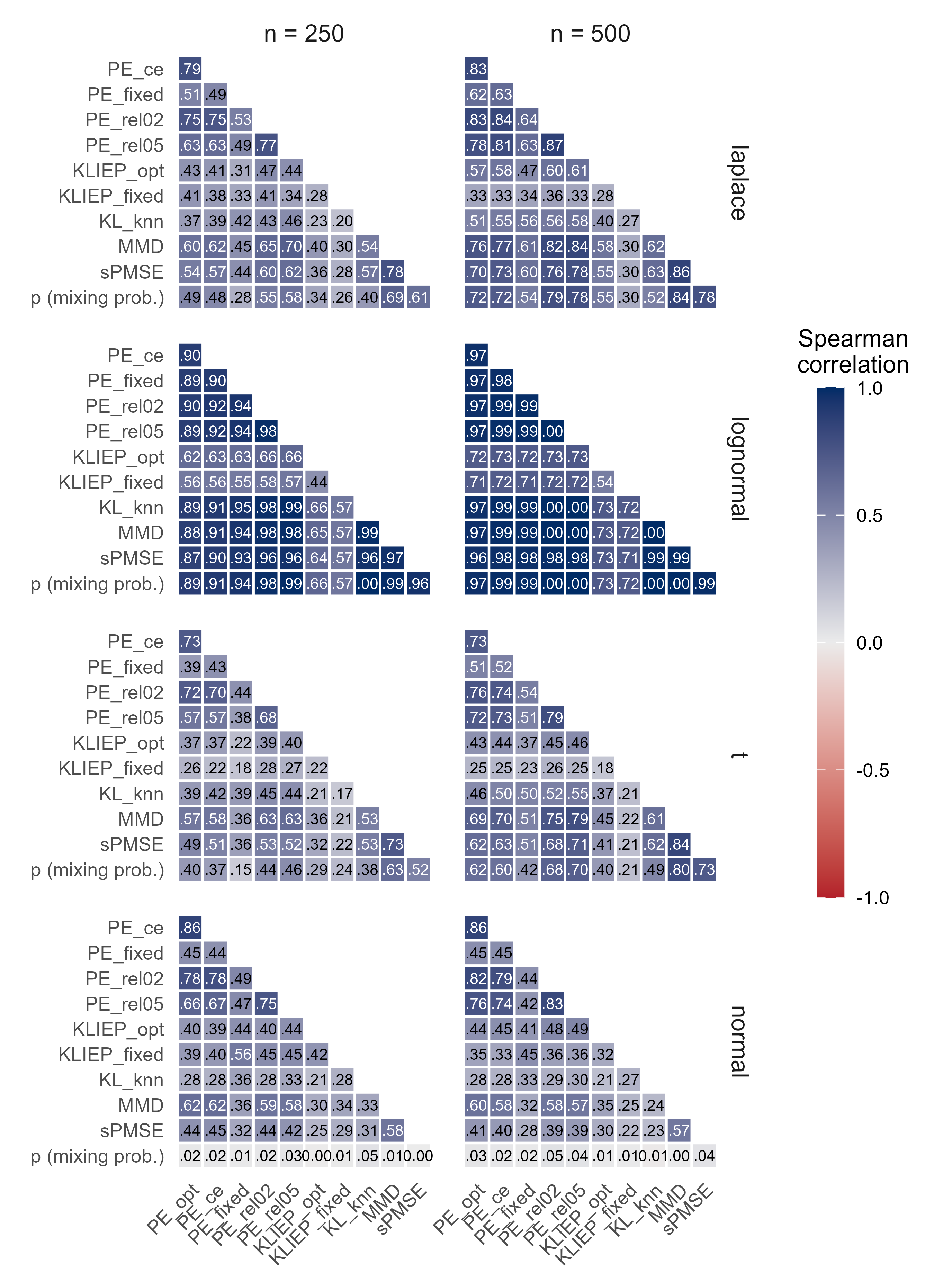}

}

\caption{\label{fig-app-stability}Spearman correlations between
divergences estimated under different density ratio model specifications
and utility statistics from alternative utility measures for each set of
synthetic data sets, averaged over 500 simulation runs.}

\end{figure}%

\linespread{2}

The correlations show that the different utility measures correlate
quite strongly among each other and with the mixture probability,
especially for the larger sample sizes. The notable exception here is
KLIEP, which performs substantially worse than the other methods (i.e.,
KLIEP has substantially lower correlations with the true mixture
probability). Moreover, all uLSIF-based measures are highly correlated
among each other, showing that the exact model specification is not
crucial for obtaining reasonable divergence estimates, although the
uLSIF model with fixed hyperparameters has slightly lower correlations
with the true density ratio models. Additionally, the relative uLSIF
variants have slightly stronger correlations with the mixing
probabilities here, while their performance was slightly worse than
plain uLSIF in the multivariate simulations. Finally, we remark that the
\(pMSE\)-ratio and the \(MMD\) statistic perform best in these
simulations, in the sense that they obtain the highest correlations with
the mixing probability.

\section{Multivariate simulations: details and additional
results}\label{sec-app-multivariate}

In this Appendix, we describe the data-generating mechanisms and
synthesis models of the multivariate simulations in more detail and
report the results of the specifications that are not shown in
Table~\ref{tbl-sim2-results} and Table~\ref{tbl-sim3-results}. Each
setting is repeated \(500\) times, with \(\nsyn = \nobs\).

\subsection{Setting 1: continuous data}\label{setting-1-continuous-data}

Let \(D_0 = D - 1\) if \(D = 5\), and \(D_0 = D - 5\) otherwise. The
first \(D_0\) variables are drawn from a multivariate normal
distribution with zero means, unit variances and all covariances equal
to \(0.5\). The remaining variables are noisy polynomials of the first
variables: \[
X_{D_0 + j} = X_j^{j+1} + \varepsilon_j, \qquad \varepsilon_j \sim \mathcal{N}\big(0, \hat{\text{Var}}(X_j^{j+1})\big),
\] with \(j = 1\) if \(D = 5\), and \(j = 1, \dots, 5\) otherwise. Here,
\(\hat{\text{Var}}(X_j^{j+1})\) is the variance of \(X_j^{j+1}\) in the
generated sample, so that roughly half of the variance of each
non-linear variable is noise.

All synthesis models are estimated on the observed data. \(M_1\) draws
each variable independently from a normal distribution with the observed
mean and variance. \(M_2\) draws all variables from a multivariate
normal distribution with the observed mean vector and covariance matrix.
\(M_3\) draws the first \(D_0\) variables from a multivariate normal
distribution with their observed mean vector and covariance matrix. The
remaining variables are then generated from the linear regression of
\(X_{D_0 + j}\) on \(X_j^{j+1}\) (with intercept) estimated on the
observed data, evaluated at the synthetic \(X_j\) and with normally
distributed residuals.

\subsection{Setting 2: mixed data}\label{setting-2-mixed-data}

The observed data consist of three categorical variables \(G_1\),
\(G_2\) and \(G_3\), with \(2\), \(4\) and \(12\) levels, and five
continuous variables \(\mathbf{X} = (X_1, \dots, X_5)^T\). We draw
\(G_1\) with \(P(G_1 = 1) = 3/4\) and \(P(G_1 = 2) = 1/4\). Given
\(G_1\), we draw \(G_2\) and \(G_3\) independently, with
\(P(G_2 \mid G_1 = 1) = (2/3, 1/4, 1/24, 1/24)\),
\(P(G_2 \mid G_1 = 2) = (1/3, 1/3, 1/9, 2/9)\),
\(P(G_3 = k \mid G_1 = 1) = k/78\) and
\(P(G_3 = k \mid G_1 = 2) = (13 - k)/78\), for \(k = 1, \dots, 12\). The
continuous variables follow \[
\mathbf{X} \mid G_1 = g, G_3 = k \sim \mathcal{N}_5(a_k \mathbf{b}, \Sigma_g),
\] with \(a_k = -1 + 2(k - 1)/11\) and
\(\mathbf{b} = (1, 0.5, 0, -0.5, -1)^\top\). Both covariance matrices
have unit variances, with all covariances equal to \(0.5\) in
\(\Sigma_1\) and \(0.6\) in \(\Sigma_2\). Hence, the categorical
variables are related through \(G_1\), the means of the continuous
variables depend on \(G_3\), and their covariances depend on \(G_1\).

All three synthesis models first generate data from this model, with
parameters estimated from the observed data: relative frequencies for
the probabilities, means within the levels of \(G_3\), and residual
covariance matrices within the levels of \(G_1\). \(M_3\) uses these
data as they are. \(M_2\) permutes the rows of the categorical variables
jointly, such that they are independent of the continuous variables but
keep their joint distribution. \(M_1\) permutes each categorical
variable separately, such that the categorical variables are also
independent of each other.

\subsection{Implementation of the utility
measures}\label{implementation-of-the-utility-measures}

For all measures except the \(pMSE\), continuous variables are
standardized with the means and standard deviations of the observed
data, and categorical variables are coded with one indicator per
category.

\begin{itemize}
\tightlist
\item
  \(PE\): uLSIF with the default settings of \texttt{densityratio}.
  These use \(200\) Gaussian kernel centers sampled from the combined
  observed and synthetic data. Leave-one-out cross-validation selects
  from \(10\) bandwidths, based on the \(0.05, \dots, 0.95\) quantiles
  of the distances between the observed records and the centers, and
  \(20\) values of \(\lambda\) between \(10^{3}\) and \(10^{-3}\).
\item
  \(PE_\text{fixed}\): uLSIF with the bandwidth based on the \(0.2\)
  quantile of these distances and \(\lambda = 1/\nobs\).
\item
  \(PE_\text{rel}\): the relative Pearson divergence from relative uLSIF
  with \(\alpha = 0.2\) (\citeproc{ref-yamada_rulsif_2013}{Yamada et al.
  2013}), otherwise with the default settings.
\item
  \(KL_\text{KLIEP}\): KLIEP with the default settings of
  \texttt{densityratio}, which select the bandwidth by five-fold
  cross-validation over the same grid. The divergence is the average log
  density ratio over the observed records.
\item
  \(KL_1\) and \(KL_{\sqrt{n}}\): \(k\)-nearest neighbor estimates with
  \(k = 1\) and \(k = \lfloor \sqrt{\nobs} \rfloor\).
\item
  \(MMD\): the unbiased estimate of the squared maximum mean discrepancy
  with a Gaussian kernel, with the bandwidth set by the default
  heuristic in \texttt{kernlab}.
\item
  \(pMSE_\text{CART}\) and \(pMSE_{\text{CART}_{cp}}\): the
  \(pMSE\)-ratio from \texttt{synthpop} based on a CART model with
  complexity parameter \(cp = 10^{-3}\) (the default) and
  \(cp = 10^{-6}\). The null expectation is estimated from \(50\)
  permutations.
\item
  \(pMSE_\text{logit}\) and \(pMSE_\text{logit, int}\): the
  \(pMSE\)-ratio based on a logistic regression with main effects, and
  with main effects and all two-way interactions. The interaction model
  is only fitted if its number of parameters does not exceed \(\nobs\).
  This excludes \((\nobs, D) \in \{(100, 25), (100, 50), (1000, 50)\}\)
  in setting 1 and \(\nobs = 100\) in setting 2.
\end{itemize}

\subsection{Additional results}\label{additional-results}

\linespread{1}

\begin{table}

\caption{\label{tbl-app-multivariate}Proportion of simulations in which
the utility measure orders the three synthetic data sets correctly, for
default uLSIF (\(PE\)), uLSIF with fixed hyperparameters
(\(PE_\text{fixed}\)), relative uLSIF with \(\alpha = 0.2\)
(\(PE_\text{rel}\)), KLIEP (\(KL_\text{KLIEP}\)), and the \(pMSE\)-ratio
based on a CART model with \(cp = 10^{-3}\) (\(pMSE_\text{CART}\)) and
\(cp = 10^{-6}\) (\(pMSE_{\text{CART}_{cp}}\)). The columns \(PE\),
\(PE_\text{fixed}\) and \(pMSE_{\text{CART}_{cp}}\) are repeated from
Table~\ref{tbl-sim2-results} and Table~\ref{tbl-sim3-results} for
comparison.\vspace{0.4cm}}

\centering{

\begin{tabularx}{\linewidth}{>{\raggedright\arraybackslash}X>{\raggedright\arraybackslash}X>{\centering\arraybackslash}p{12mm}>{\centering\arraybackslash}p{12mm}>{\centering\arraybackslash}p{12mm}>{\centering\arraybackslash}p{17mm}>{\centering\arraybackslash}p{17mm}>{\raggedright\arraybackslash}X}
\toprule
$n$ & $D$ & $PE$ & $PE_{\text{fixed}}$ & $PE_{\text{rel}}$ & $KL_{\text{KLIEP}}$ & $pMSE_{\text{CART}}$ & $pMSE_{\text{CART}_{cp}}$\\
\midrule
\addlinespace[0.3em]
\multicolumn{8}{l}{\textbf{Setting 1: continuous data}}\\
\hspace{1em}100 & 5 & 0.65 & 0.84 & 0.63 & 0.88 & 0.34 & 0.35\\
\hspace{1em}100 & 25 & 0.63 & 0.68 & 0.61 & 0.44 & 0.32 & 0.32\\
\hspace{1em}100 & 50 & 0.54 & 0.59 & 0.52 & 0.44 & 0.37 & 0.37\\
\hspace{1em}1000 & 5 & 1.00 & 1.00 & 1.00 & 1.00 & 0.87 & 0.95\\
\hspace{1em}1000 & 25 & 1.00 & 1.00 & 1.00 & 0.94 & 0.89 & 0.92\\
\hspace{1em}1000 & 50 & 0.98 & 1.00 & 0.95 & 0.91 & 0.83 & 0.87\\
\hspace{1em}2000 & 5 & 1.00 & 1.00 & 1.00 & 1.00 & 0.90 & 0.97\\
\hspace{1em}2000 & 25 & 1.00 & 1.00 & 1.00 & 0.99 & 0.85 & 0.99\\
\hspace{1em}2000 & 50 & 1.00 & 1.00 & 1.00 & 0.96 & 0.78 & 0.96\\
\addlinespace[0.3em]
\multicolumn{8}{l}{\textbf{Setting 2: mixed data}}\\
\hspace{1em}100 & 8 & 0.14 & 0.57 & 0.15 & 0.33 & 0.24 & 0.24\\
\hspace{1em}1000 & 8 & 0.99 & 0.99 & 0.98 & 0.70 & 0.58 & 0.59\\
\hspace{1em}2000 & 8 & 1.00 & 1.00 & 1.00 & 0.78 & 0.71 & 0.71\\
\bottomrule
\end{tabularx}

}

\end{table}%

\linespread{2}

Table~\ref{tbl-app-multivariate} shows that relative uLSIF is
essentially indistinguishable from uLSIF. KLIEP performs substantially
worse than uLSIF, except when \(n = 100\) and the number of variables is
small. The \(pMSE\)-ratio with default complexity parameter performs
poorer than the \(pMSE\)-ratio with smaller complexity parameter, except
when the sample size is small, in which case both perform comparable.
The uLSIF-estimators perform substantially better than both KLIEP
(except for \(n = 100\) and \(D = 5\) in Setting 1 and \(n = 100\) in
Setting 2) and both \(pMSE\)-ratio models (except \(n = 100\) in Setting
2).

\end{document}